# Human-like Cognitive Generalization for Large Models via Brain-in-the-loop Supervision

Jiaxuan Chen[1,2#], Yu Qi[2,3#*], Yueming Wang[1], Gang Pan[1,2*]

[1]College of Computer Science and Technology, Zhejiang University, China

[2]State Key Lab of Brain-Machine Intelligence, Zhejiang University, China

[3]Affiliated Mental Health Center and Hangzhou Seventh People's Hospital, MOE Frontier Science Center for Brain Science and Brain-machine Integration, Zhejiang University, Hangzhou, China.

# These authors contributed equally to this work.

* Corresponding Authors.

Correspondence and requests for materials should be addressed to:

Yu Qi (qiyu@zju.edu.cn)

**Abstract**

Recent advancements in deep neural networks (DNNs), particularly large-scale language models, have demonstrated remarkable capabilities in image and natural language understanding. Although scaling up model parameters with increasing volume of training data has progressively improved DNN capabilities, achieving complex cognitive abilities—such as understanding abstract concepts, reasoning, and adapting to novel scenarios, which are intrinsic to human cognition—remains a major challenge. In this study, we show that brain-in-the-loop supervised learning, utilizing a small set of brain signals, can effectively transfer human conceptual structures to DNNs, significantly enhancing their comprehension of abstract and even unseen concepts. Experimental results further indicate that the enhanced cognitive capabilities lead to substantial performance gains in challenging tasks, including few-shot/zero-shot learning and out-of-distribution recognition, while also yielding highly interpretable concept representations. These findings highlight that human-in-the-loop supervision can effectively augment the complex cognitive abilities of large models, offering a promising pathway toward developing more human-like cognitive abilities in artificial systems.

## Introduction

Humans live in a world full of ever-changing objects, which requires us to constantly learn new concepts from a variety of sensory information and systematically structure our experiential knowledge of the physical world[1,2]. This ability helps us organize our knowledge and make sense of the world, allowing us to recognize new things and adapt to new situations. For years, artificial intelligence (AI) researchers have tried to build systems that understand the world like humans do[3–7].

Deep neural networks (DNNs) have made great progress, especially with the help of large-scale datasets and large models (LMs)[8–10], while they still struggle to match human-like cognition[11]. Scaling up data and model size has primarily improved performance in well-defined tasks like classification, object detection, and content generation[12,13]. However, when it comes to human-like cognitive abilities—particularly the understanding and generalization of novel concepts—current models still fall short[1,11,14]. Moreover, increasing model size does not consistently enhance these capabilities, and in some cases, performance even plateaus or degrades[11,15], highlighting a fundamental bottleneck in achieving human-like cognition.

Here, we proposed to enhance the cognitive ability of DNNs by leveraging the structural organization of the human brain as a supervisory signal. Our initial observations revealed that while increasing the parameter size of DNNs improved their understanding of concrete concepts (Fig. 1a), it had little effect on abstract concepts (Fig. 1b, c). This limitation stems from DNNs' inability to capture the internal relationships among concepts and their lack of abstraction and conceptualization capabilities (Fig. 1d, e). Motivated by this insight, we developed a brain-in-the-loop supervision framework to directly transfer the brain's cognitive structure to DNNs. Remarkably, we found that aligning the structural representations of DNNs with the brain using only a small set of objects significantly enhanced their ability to understand abstract concepts (Fig .1b, c), with the learned cognitive structures generalizing well to unseen concepts. This process led to the emergence of human-like conceptual hierarchies in DNNs, closely resembling WordNet[16]—even without explicit constraints to match it. The proposed framework consistently improved performance across a range of challenging tasks, including few-shot/zero-shot learning[17,18] and out-of-distribution recognition[19], while also producing highly interpretable concept representations. These results were robust across diverse DNN architectures, including SimCLR[20], CLIP[9], and DINOv2[8]. Our findings suggest a promising new direction for developing AI systems that think more like humans, moving beyond the limitations of current approaches that rely solely on scaling data and model complexity.

## Results

### Increasing parameter sizes of enhances DNN's understanding of concrete but not abstract concepts

To assess the concept understanding capabilities of DNNs, we first conducted one-shot learning tasks for both concrete and abstract concepts. The concrete concepts comprised 50 classes, including objects such as *sock* and *swan*, while the abstract concepts consisted of 8 higher-level categories, such as *clothing* and *bird* (Supplementary Fig. S1). We employed the one-shot learning task, where for each class, one image was randomly selected for training, and the remaining images were used for test (see Supplementary for details). The abstract concept task posed a significantly greater challenge, as it required models to generalize to unseen objects within a broader category. For instance, the model had to infer that an *owl* belongs to the class *bird* when provided with a *swan* as the only example.

We evaluated several DNN architectures, including SimCLR[20], DINOv2[8], and CLIP[9], across varying parameter scales (see Supplementary for details). For concrete concepts, we observed a consistent improvement in one-shot classification performance as the number of parameters increased (Fig. 1a). In contrast, for abstract concepts, increasing model parameters yielded only marginal improvements or even led to performance degradation (Fig. 1b, c). For instance, with the DINOv2 model, performance on concrete concepts significantly improved from 74.94% to 85.87% ($P = 1.78 \times 10^{-76}$, two-tailed paired $t$-test) as trainable parameters increased from 22.06 million to 304.37 million. However, for abstract concepts, the performance gain from scaling parameters was negligible or even negative, with performance dropping significantly from 54.37% to 52.82% as parameters increased from 22.06 million to 304.37 million ($P = 6.66 \times 10^{-4}$, two-tailed paired $t$-test). Similar trends were observed across SimCLR, and CLIP models. These results revealed that while increasing the parameter size of DNNs improved their performance on concrete concepts, it did not substantially enhance their ability to understand abstract concepts.

To investigate the underlying reasons for the performance disparity between concrete and abstract concepts, we analyzed the representational dissimilarity matrices (RDMs) to visualize the conceptual relationships (representational distances) learned by each model. We first examined the brain's RDM as a reference for a well-structured conceptual framework (Fig. 1d). We observed that the brain's RDM revealed two distinct clusters (Fig. 1d, see Supplementary for details) corresponding to living beings (red square) and non-living objects (purple square), reflecting a robust semantic understanding of the concepts. When comparing the RDMs of DNNs, only the CLIP model exhibited partial alignment with the brain's RDM, showing a discernible cluster for living beings (Fig. 1e). In contrast, the other models, while demonstrating strong discrimination of concrete concepts (evidenced by superior classification capabilities), failed to capture the broader semantic clusters of living and non-living entities. This suggests that without a meaningful understanding of the relationships between concrete concepts, DNNs struggle to generalize to abstract concepts, highlighting a critical limitation in their cognitive capabilities.

## Brain-in-the-loop supervision enables DNNs to learn human-like conceptual representations

Previous analysis indicated that the inability to capture internal relationships between concepts might be a key factor limiting DNNs' understanding of abstract concepts—a capability inherently embedded in human cognition[2]. Instead of further scaling up model parameters, we proposed a brain-in-the-loop supervision framework to transfer human conceptual structures to DNNs.

We hypothesized that the conceptual structure of objects could be represent with a graph, and both the pretrained LMs and human brain have formed their own conceptual structures based on learning. Then given a set of objects, we can align the LMs' conceptual graph to human brain's with a graph matching[21] approach. Here we built an iterative optimization framework (Fig. 2a) to reshape the conceptual structure of LMs through two main steps that performed in an alternative manner: 1) seeking the optimal correspondences between LMs and the human brain representations; 2) updating representations to maximize structural similarity (Fig. 2b). To this end, our architecture consists of three learnable modules, i.e., a fMRI encoder, an image encoder, and a graph neural network[22]. By differentiable design, gradients can be backpropagated from the matching layers to the encoders during training procedure (see Methods for details).

While the number of concepts is large in the real world, we found that by aligning the conceptual graph with a small set of objects, the alignment was well generated to unseen concepts. Here, using the CLIP model as an example, we performed alignment between the brain and the CLIP model across 150 concrete object categories, and evaluated the distances between the representations of brain and DNN with 50 unseen classes (see Methods). As shown in Fig. 2c, as the brain-in-the-loop supervision process evolves, the distance between brain and DNN representations decreased consistently for both the concepts in the training concepts and unseen concepts in mostly linear processes ($R^2 = 0.86$, Pearson's $r = -0.93$ for training, and $R^2 = 0.83$, Pearson's $r = -0.91$ for unseen), revealing good generalization to unseen objects. The RDM of the 50 unseen classes also demonstrated similarity to brain's RDM (Supplementary Fig. S2a), with a better conceptual structure of objects (Supplementary Fig. S2b, c). These results suggested that brain-in-the-loop supervision learning approach effectively transferred human's conceptual structures to DNNs.

**Enhancement of abstract concept understanding**

We next evaluated the model's ability to understand abstract concepts with and without brain-in-the-loop supervision. As illustrated in Fig. 1a and b (Bottom), models trained with brain-in-the-loop supervision demonstrated significant improvements in one-shot learning tasks for abstract concepts ($P < 0.0001$, two-tailed paired $t$-test, n = 100).

To further assess the understanding of abstract concepts at different levels of abstraction, we employed the Silhouette Coefficient (SC)[23] to measure the quality of representations on 50 unseen classes (n = 66,067). A higher SC indicates that objects within the same concept have similar representations, while objects from different concepts are well-separated, reflecting a better understanding of conceptual relationships. As shown in Fig. 3a, without brain-in-the-loop supervision, models exhibited strong representation ability for concrete concepts (SC = 0.231 for the 50 leaf nodes) but showed a marked decline in performance for middle-level (SC = 0.144 for the 8 middle-level nodes) and high-level concepts (SC = 0.044 for the 3 high-level nodes). This suggests that while models excel at understanding concrete concepts, they struggle with abstract ones.

In contrast, with brain-in-the-loop supervision, concept representation improved significantly across all three abstraction levels (Fig. 3a, $P < 0.0001$, two-tailed paired $t$-test), particularly for abstract concepts. The SC increased from 0.144 to 0.239 for middle-level concepts and from 0.044 to 0.127 for high-level concepts. Notably, we observed a nearly linear correlation between the degree of structural alignment (measured by Gromov-Wasserstein (GW) distance[24]) and the SC score (Fig. 3b). This strongly suggests that the enhanced conceptual understanding stems from the alignment of the model's representational structure with the brain's conceptual framework. Further improvements could be achieved by pushing the models' representational structures to be more consistent with the brain. Together, these findings demonstrate that brain-in-the-loop supervision effectively enhances models' capacity to conceptualize abstract categories.

**Emergence of human-like conceptual hierarchy**

We further assessed the similarity between the conceptual hierarchy structures of humans and models using the 50 unseen concepts. To this end, we utilized WordNet[16], a comprehensive lexical database that encodes conceptual-semantic relationships, as a representation of human conceptual hierarchy (Fig. 4a). We compared this structure of the models with the 50 unseen concepts. Our analysis revealed that models trained with brain-in-the-loop supervision exhibited concept similarity matrices (CSMs) highly aligned with WordNet, whereas models without such supervision showed no such alignment (Fig. 4b; see Supplementary Tab. 1 for the row/column labels of CSMs). These findings were consistent across different DNN architectures and brain signals from multiple subjects (Fig. 4b-d and Supplementary Fig. S3).

Additionally, we found that brain-in-the-loop supervision enabled models to exhibit human-like judgments in concept similarity tasks. Specifically, we employed a triplet odd-one-out task, where participants (human or DNN model) were presented with three images and asked to select the least related one (Fig. 4e). Human judgment data were sourced from the THINGS database[25], which contains 4.7 million triplet odd-one-out judgments across 1,854 objects from 12,340 participants. We compared the similarity between human and model responses (see Supplementary for details). Results showed that brain-in-the-loop supervision significantly improved the alignment between human judgments and model behaviors (Fig. 4f), with accuracy increasing from 43.74% to 49.94% ($P = 7.56 \times 10^{-97}$, two-tailed paired $t$-test).

Together, these results demonstrate that brain-in-the-loop supervision enables models to develop human-like conceptual hierarchy structures, significantly enhancing their ability to conceptualize abstract categories.

**Human-like conceptual hierarchy enhances performance in complex cognitive tasks**

We demonstrated that models endowed with human-like conceptual hierarchy structures achieve significant performance improvements across a variety of complex cognitive tasks.

**One-shot learning (OSL).** We first evaluated the one-shot learning capabilities of the models at different abstraction levels. As was mentioned, a central challenge in one-shot classification, is to rapidly understand the concepts from only one example[26]. The problem can be more difficult when the concepts become more abstract (Fig. 5a, see Supplementary). When there are only two abstract concepts of living thing and non-living thing, CLIP-base with brain-in-the-loop supervision achieved an average improvement of 20.5 ± 0.01% (mean ± s.d., averaged across S1-S3, $P < 0.0001$ for all subjects, two-tailed paired *t*-test) in OSL performance, outperforming baseline models with 4.9 times more parameters (Fig. 5b). Brain-in-the-loop supervision achieved consistent performance gains across different DNN architectures (CLIP[9], SimCLR[20] and DINOv2[8]) and supervision using brain signals from different subjects (Fig. 5b and Supplementary Fig. S4). Moreover, we found a nearly linear correlation between the degree of structural alignment (measured by GW distance[24]) and the one-shot classification accuracy, consistent across supervision using brain signals from different subjects (Supplementary Fig. S5).

**Generalized out-of-distribution (OOD) recognition.** Next, we examined how brain-in-the-loop supervision enhanced the robustness of generalized OOD recognition (Fig. 5c, see Supplementary). We defined a series of OOD recognition tasks (n = 31, Supplementary Tab. 2), which assess the conceptual extrapolation ability of DNNs from minimal category supervision, such as determining whether unseen categories (e.g., "bowling" and "carriage") should be classified as a broader category (e.g., "non-animal"). These tasks require models to capture high-level semantic structures that align with the hierarchical structure of human conceptual knowledge. As shown in Fig. 5d, brain-in-the-loop supervision significantly improved OOD robustness, yielding performance increases of 11.5 ± 0.2%, 13.6 ± 0.9%, 9.1 ± 0.7% (mean ± s.d., averaged across 3 subjects and 31 tasks, $P < 0.0001$ for all subjects, two-tailed paired *t*-test) for CLIP-base, SimCLR-base and DINOv2-base, respectively, without the need for extensive supervision labels. Similar improvement trends can be found in different architectures across varying parameter scales and supervision using brain signals from different subjects (Fig. 5d).

**Brain-to-image and image-to-image retrieval.** To intuitively assess the quality of the learned representations, we evaluated the effectiveness of brain-in-the-loop supervision for both brain-to-image (Fig. 5e, see Supplementary) and image-to-image (Fig. 5g, see Supplementary) retrieval tasks. Brain-to-image retrieval, a common benchmark for neural decoding models[27–29], evaluates the ability to reconstruct perceived stimuli (n = 50) from brain signals. Our approach consistently improved decoding performance, achieving an average increase of 7.34%–26.67% in 50-way retrieval across all subjects compared to state-of-the-art decoding methods[29–31] (Fig. 5f). Qualitative analysis of brain-to-image retrieval on a subset of ImageNet-21K[32] (n = 66,067, Supplementary Fig. S6a-c) indicated the concept representations of DNNs can be predicted from fMRI patterns derived using our brain-in-the-loop supervision framework. For example, the fMRI pattern evoked from an image of "a man *lying* on a hammock" can be the image of "a child *lying* on the baby carriage" (the 5$^{th}$ row, Supplementary Fig. S6a). In the case of "harp" (the 6$^{th}$ row, Supplementary Fig. S6a), although the top retrievals fall into the wrong concrete categories, are all related to "musical instruments". Similar results can be found in most instances across different subjects (Supplementary Fig. S6b, c).

In image-to-image retrieval tasks (Fig. 5g, h), brain-in-the-loop supervision enabled models to retrieve semantically similar objects despite significant visual differences, even when these objects belonged to different predefined categories. For example, an image of a "beer mug" successfully retrieved images sharing attributes like "cup" or "jug" from distinct categories (Fig. 5h, top row). Similar results were observed across most instances, highlighting the model's enhanced ability to capture semantic relationships beyond superficial visual features. Collectively, these findings demonstrate that human-like conceptual hierarchy structures, facilitated by brain-in-the-loop supervision, significantly enhance model performance in complex cognitive tasks.

**Cognitive-level consistency of the learned conceptual manifold**

To evaluate whether the conceptual manifold, shaped by brain-in-the-loop supervision, maintains cognitive-level consistency for both seen and unseen points, we trained a GPT-2-based[33] text decoder on the COCO dataset[34]. This decoder reconstructs the semantics of a given representation as a sequence of words, enabling a detailed analysis of the manifold's properties (see Supplementary for details).

**Local semantic consistency.** We first assessed the local semantic consistency of the manifold, which measures whether similar concepts are clustered together in the representation space. To this end, we designed an out-of-distribution semantic coherence test (see Supplementary for details), evaluating the semantic consistency and similarity between newly generated samples and their nearest real neighbors in PCA space (Fig. 6a). As shown in Fig. 6b, we found that the 2-D PCA points sampled from different conceptual regions show strong semantic correlations with the real samples. For example, we can reconstruct the semantics of *T-shirt* and *tie* within the "clothing" region and generate the semantics of *truck* near the region associated with "vehicles". In contrast, models without brain supervision produced semantically meaningless reconstructions (Supplementary Fig. S7a). By using 8 abstract categories as ground-truth, we found that brain-in-the-loop supervision significantly enhanced local semantic coherence, with semantic consistency accuracy improving by 15.9% and semantic similarity scores increasing by 6.2% on average compared to the model without supervision ($P < 0.0001$, two-tailed paired $t$-test, Fig. 6c).

**Intersections between different concepts.** We further investigated whether intersections between different concepts yield meaningful reconstructions. By generating new points through interpolation between two real samples from distinct concepts (Fig. 6d), we observed that the reconstructed semantics aligned well with human intuition, exhibiting smooth and realistic transitions (Fig. 6e). For instance, transitioning between "iPod" and "video recorder" resulted in a logical semantic shift from "television" to "video recorder" (Fig. 6e, left), while interpolating between vehicles and ungulate animals produced the concept of "horse" (Fig. 6e, right). Qualitative analysis indicates most of the key features of these concepts can be effectively captured with a limited number of principal components, which is also reflected in the reconstruction error and explained variance ratio (Supplementary Fig. S7b). These findings suggest that brain-in-the-loop supervision ensures the conceptual manifold is semantically continuous and smooth, enhancing the model's cognitive generalization capabilities.

**Concept summarization and arithmetic.** Finally, we evaluated the conceptual manifold using two highly challenging tasks: concept summarization (Fig. 6f) and concept arithmetic (Fig. 6h, see Supplementary for details). For concept summarization, the model was tasked with generating a sentence to describe multiple input images. As shown in Fig. 6g, the regulated manifold enabled the model to infer shared behavioral details and the number of objects. For example, given a set of images depicting different types of animals lying on grass, the decoder with brain-in-the-loop supervision was able to reconstruct the sentence of "animal laying on a grassy area". Given a set of images, each depicting a close-up of a pair of different animals standing on grass, the decoder successfully reconstructed the text "two animals standing in a grassy area".

For concept arithmetic (Fig. 6h), we tested whether brain-in-the-loop supervision could steer generated captions toward desired conceptual directions through subtraction and addition operations. For example, subtracting "motorcycle" and adding "horse" shifted the semantics from "a man riding a motorcycle" to "a man riding a horse" (Fig. 6i, first case). Similarly, removing "dog" from "a dog playing with a frisbee" and adding "a young girl" yielded "a young girl is playing with a frisbee" (Fig. 6i, second case). In the third case, subtracting "banana" and adding "cell phone" transformed "a young girl holding a banana" into "a young girl holding a cell phone."

These results demonstrate that brain-in-the-loop supervision produces a low-dimensional conceptual manifold that aligns with human judgments, enabling accurate and intuitive descriptions of objects along semantically meaningful dimensions. This advancement highlights the potential of brain-guided approaches in creating models with enhanced cognitive generalization and interpretability.

## Discussion

Our study demonstrates that incorporating brain-derived structural representations as a supervisory signal can significantly enhance the cognitive capabilities of DNNs, particularly in abstract concept understanding and generalization. Unlike traditional scaling approaches—which prioritize increasing data volume or model size—the brain-in-the-loop framework enables DNNs to develop human-like conceptual hierarchies, even when trained on very limited examples (150 classes). This suggests that structural alignment with biological cognition may be a crucial yet underappreciated factor in bridging the gap between artificial and human intelligence.

Our results support that human-like cognition relies not merely on statistical learning from large data but also on structured knowledge organization. While expending training dataset and model parameters has improved DNNs' performance in recognizing concrete concepts[10], such abilities remain rooted in statistical correlations rather than conceptual abstraction[11,15]. Human cognition benefits from lifelong, interactive learning—a process that integrates continuous sensory input, active exploration, and real-world engagement to refine conceptual relationships. These mechanisms, which are inherently difficult to replicate in DNN training paradigms, underscore the limitations of purely scale-driven approaches in achieving human-level cognitive flexibility.

Prior work has sought to enhance DNNs' cognitive abilities by aligning their behavior with humans in narrow domains[35,36], such as conceptual combination skills[1] and human triplet odd-one-out responses[37]. While valuable, these efforts often focus on task-specific optimizations rather than generalizable cognitive structures. Our framework diverges by leveraging the brain's representational geometry as a universal supervisory signal. Remarkably, this approach elicited emergent conceptual hierarchies in DNNs resembling human systems like WordNet—without explicit hierarchical constraints—and generalized to few-shot and zero-shot learning scenarios. These results suggest that the brain's intrinsic representational structure encodes abstractions transferable to artificial systems, offering a pathway to more interpretable and robust generalization.

Further, our work highlights the potential of neurocognitive principles to advance AI systems. The brain-in-the-loop framework not only improves model performance but also yields interpretable representations—an essential step toward AI that reasons like humans. Future efforts could integrate this approach with large-scale models, combining the strengths of data-driven learning with biologically grounded structural priors.

A key limitation of our framework is its reliance on subject-specific neural data, which may hinder scalability due to individual differences in brain responses. Future research should explore methods to distill population-invariant conceptual structures from large-scale neuroimaging datasets, enabling broader applicability. Additionally, while our study focused on static visual and semantic concepts, extending this approach to dynamic, multimodal environments (e.g., temporal reasoning or embodied interaction) could further validate its universality.

To summarize, our findings demonstrate that brain-aligned supervision enables DNNs to achieve human-like cognitive generalization, offering a viable alternative to scale-centric AI development. This paradigm shift—from "bigger is better" to "structured is smarter"—holds significant promise for building more intelligent, adaptable, and interpretable artificial systems, narrowing the gap between machine and human intelligence.

# Figures

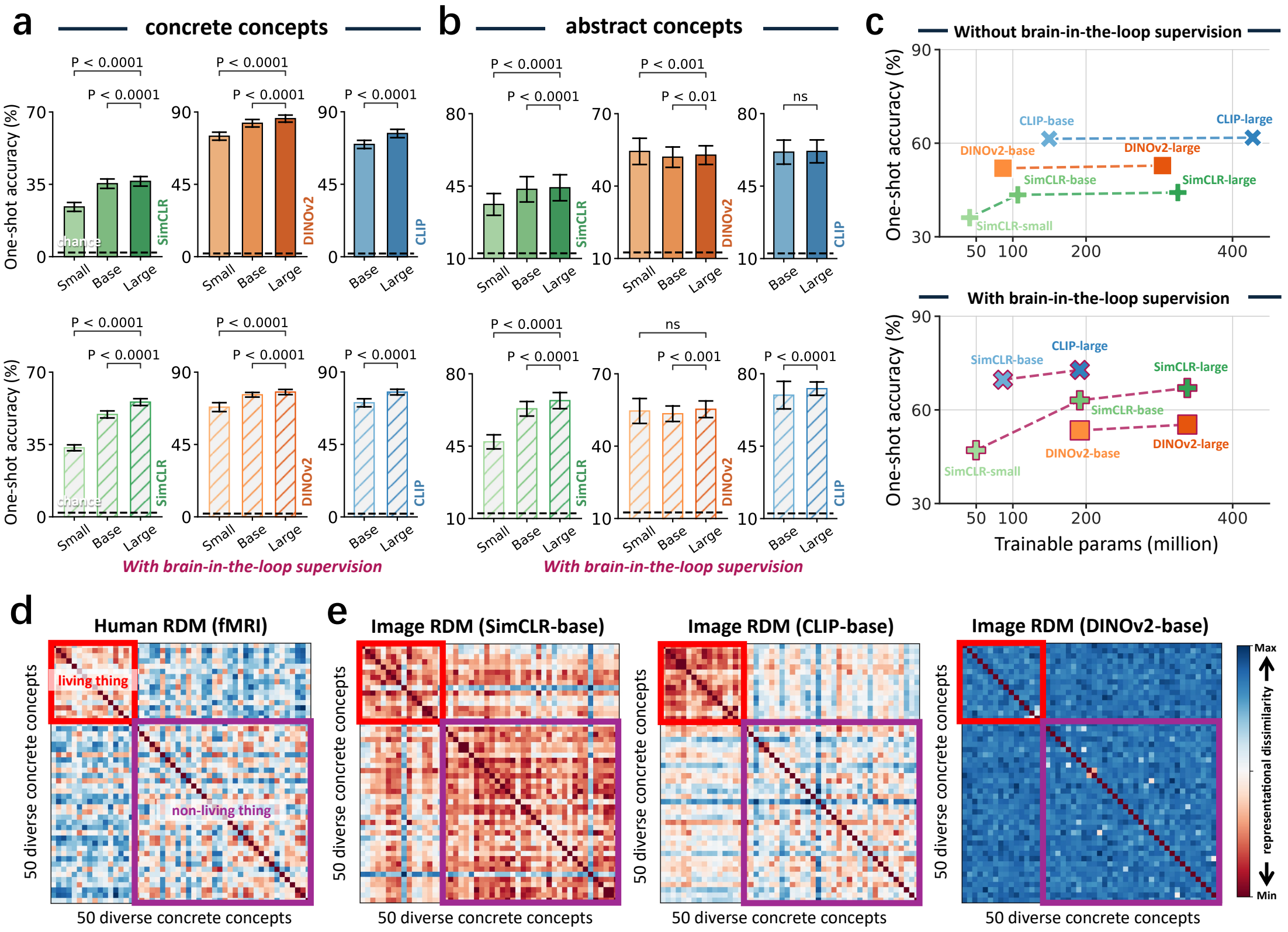

**Fig. 1 | Concept understanding capabilities of DNNs and brain. a-c.** One-shot classification performance using DNNs without (above) and with (below) brain-in-the-loop supervision, on concrete concepts (**a**) and abstract concepts (**b**). Without brain-in-the-loop supervision, increasing parameter sizes in DNNs generally improved the performance on concrete concepts (**a**, n = 50)**,** but not abstract concepts (**b, c**, n = 8). With brain-in-the-loop supervision, the DNNs obtained significant improvement in abstract concept understanding (**b, c**). **d, e.** Representational dissimilarity matrices (RDMs) of human visual cortex (**d**) and DNNs (**e**). In **d**, the RDMs were computed by using fMRI data that was recorded while subjects were viewing stimulus images (50 concrete concepts, averaged across three subjects S1-S3, see Supplementary). In **e**, RDMs of different DNN architectures were computed by using the latent embeddings of the stimuli images. In each RDM, the red square denotes concepts belonging to living thing, and the purple square contains all non-living objects (see Supplementary).

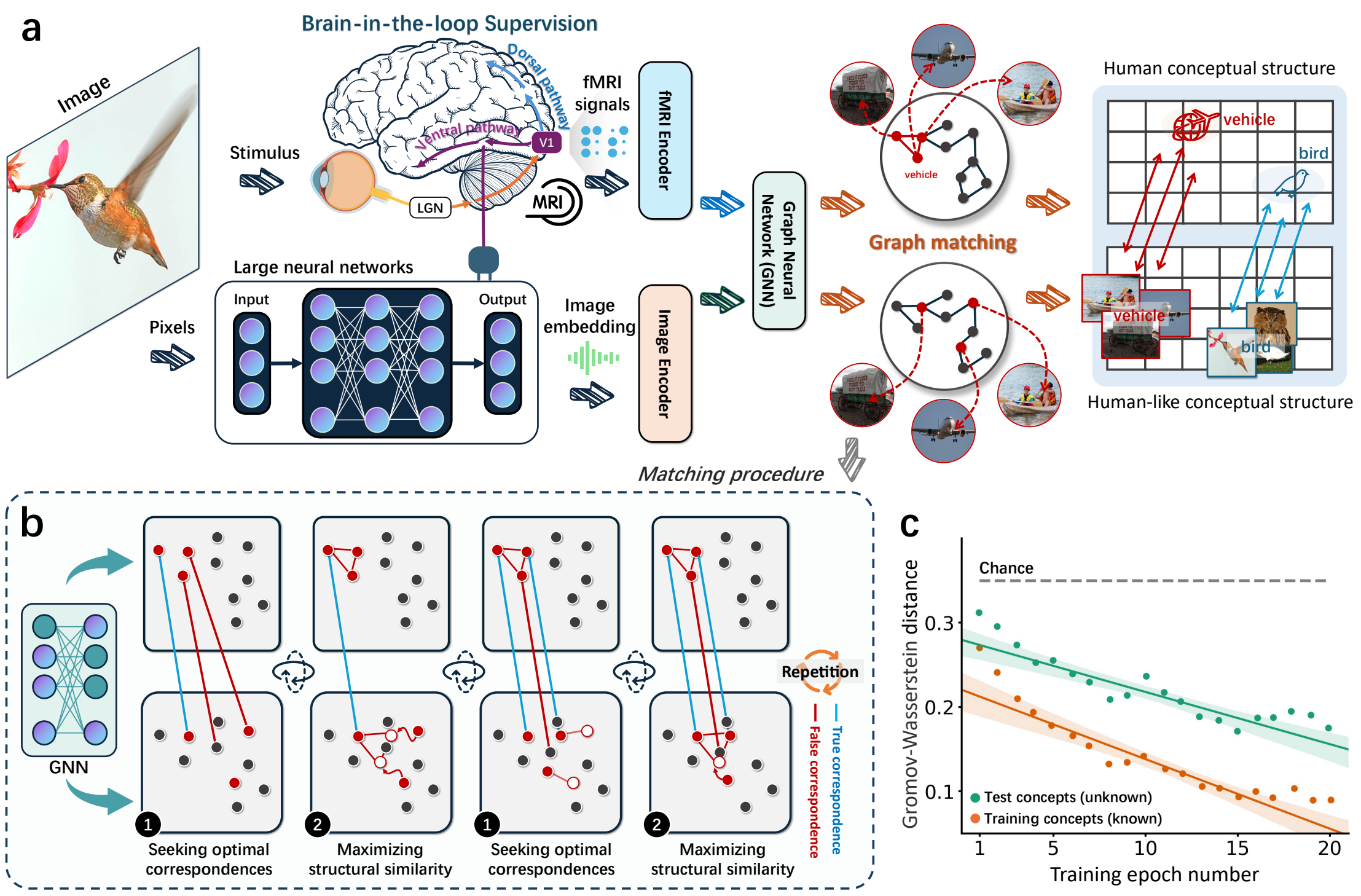

**Fig. 2 | The brain-in-the-loop supervision approach. a.** Overview of brain-in-the-loop supervised learning framework. Given the same stimuli to the brain and DNN model, the framework aligns the representation of the DNN (extracted by the Image encoder) to the representation of the brain (extracted by the fMRI encoder) via a graph matching approach. **b.** The graph matching process. The optimization procedure includes two main alternating learning steps: 1) seeking a set of putative optimal correspondences between fMRI and image embeddings with minimum Wasserstein distance; 2) updating node representations to maximize structural similarity (see Methods). **c.** The structure similarity between DNN and brain in the training process. The brain-in-the-loop supervision leads to improved conceptual structure similarity (measured in GW distance) between the brain and DNNs (CLIP-base model) for both training concepts (red dots; Pearson's $r$ = -0.93, n = 150), and unseen concepts (green dots; Pearson's $r$ = -0.91, n = 50).

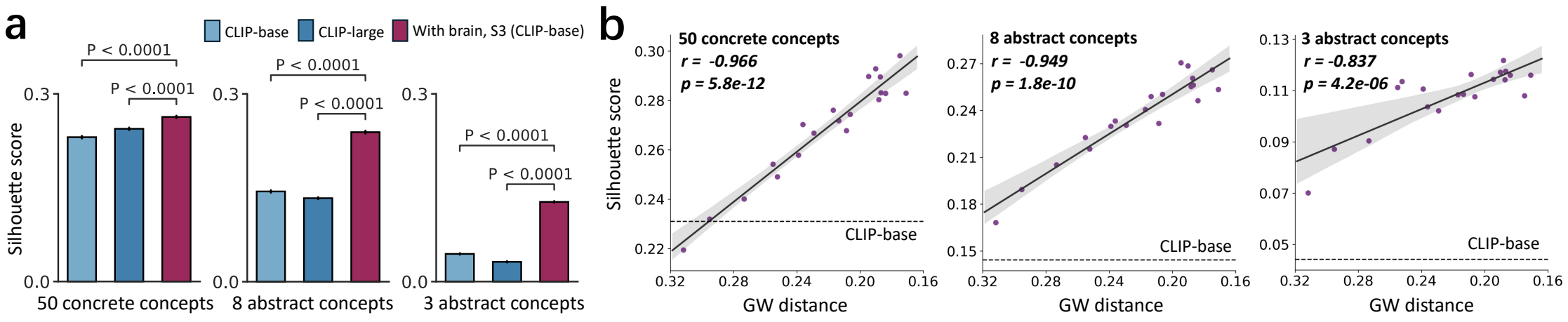

**Fig. 3 | Brain-in-the-loop supervision enhances the ability to understand abstract concepts for DNNs. a.** Silhouette coefficient (SC) scores were computed for DNNs with and without brain-in-the-loop supervision (n = 50 independent runs; 20% test samples (13, 213) randomly selected per run). Higher SC values indicate more distinct clustering of representations, which reflect better understanding of concepts. The brain-in-the-loop supervision process significantly improved the SC scores for DNNs, especially with abstract concepts (CLIP-base vs CLIP-base with brain-in-the-loop supervision: $P = 1.44 \times 10^{-75}$ for 50 concrete concepts, $P = 4.32 \times 10^{-89}$ for 8 abstract concepts, $P = 3.07 \times 10^{-93}$ for 3 abstract concepts, two-tailed paired *t*-test). **b.** Relationship between SC scores and GW distance (n = 20 models were plotted). A strong negative correlation was observed (Pearson's $r = -0.966$, $p = 5.8 \times 10^{-12}$ for 50 concrete concepts, Pearson's $r = -0.949$, $p = 1.8 \times 10^{-10}$ for 8 concrete concepts, Pearson's $r = -0.837$, $p = 4.2 \times 10^{-6}$ for 3 concrete concepts), indicating enhanced conceptual understanding as the DNN's representation aligned more closely with neural representations in the brain. Each SC score was calculated using all test samples.

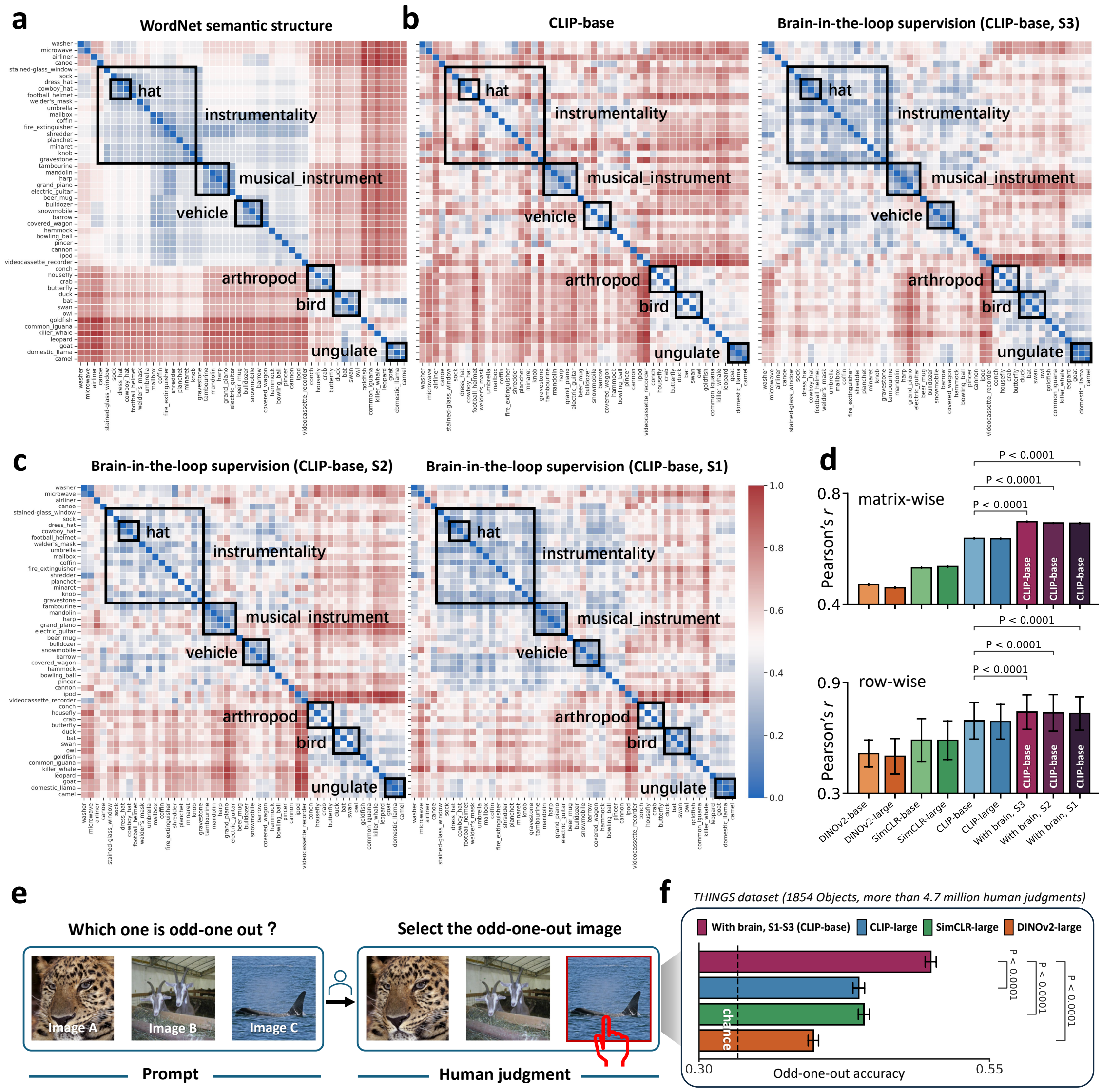

**Fig. 4 | Human-like conceptual hierarchy emerges in DNNs after brain-in-the-loop supervision. a.** Conceptual similarity matrix (CSM) of WordNet. Values in the CSM were calculated based on the pairwise shortest path between WordNet synsets (see Supplementary). The rows of the CSM for WordNet were ordered by the average-linkage hierarchical clustering algorithm to expose the semantic structure patterns (see Supplementary Tab. 1 for the sorted row/column labels). **b.** CSMs of DNNs with and without brain-in-the-loop supervision for 50 unseen concepts. After supervision, the DNN's CSM showed increased similarity to WordNet's structure (matrix-wise Pearson's $r$ = 0.70 for CLIP-base with brain-in-the-loop supervision, matrix-wise Pearson's $r$ = 0.64 for CLIP-base), demonstrating improved acquisition of human-like conceptual representations that generalized effectively to novel concepts. **c.** The CSMs were consistent using different individuals' brain signals for brain-in-the-loop supervision. See Supplementary Fig. S3 for more examples. **d.** Evaluation of similarity between CSMs of WordNet and DNNs using the Pearson correlation $r$. With supervision, matrix-wise and row-wise Pearson correlations increased from 0.64 and 0.70 (CLIP-base) to [0.69, 0.69, 0.70] and [0.73, 0.74, 0.74] (S1-S3), respectively ($P < 0.0001$ for both, two-tailed paired $t$-test). **e.** Human triplet odd-one-out judgment experiment. Participants (human or DNN) were asked to select

the one that is least similar to the other two images. **f.** Results of the triplet odd-one-out judgment task. Brain-in-the-loop supervision significantly increased the response consistency with human judgments (from 43.74% to 49.94% (averaged across S1-S3), chance level = 33.3%; $P = 7.56 \times 10^{-97}$, two-tailed paired $t$-test; see Supplementary for details).

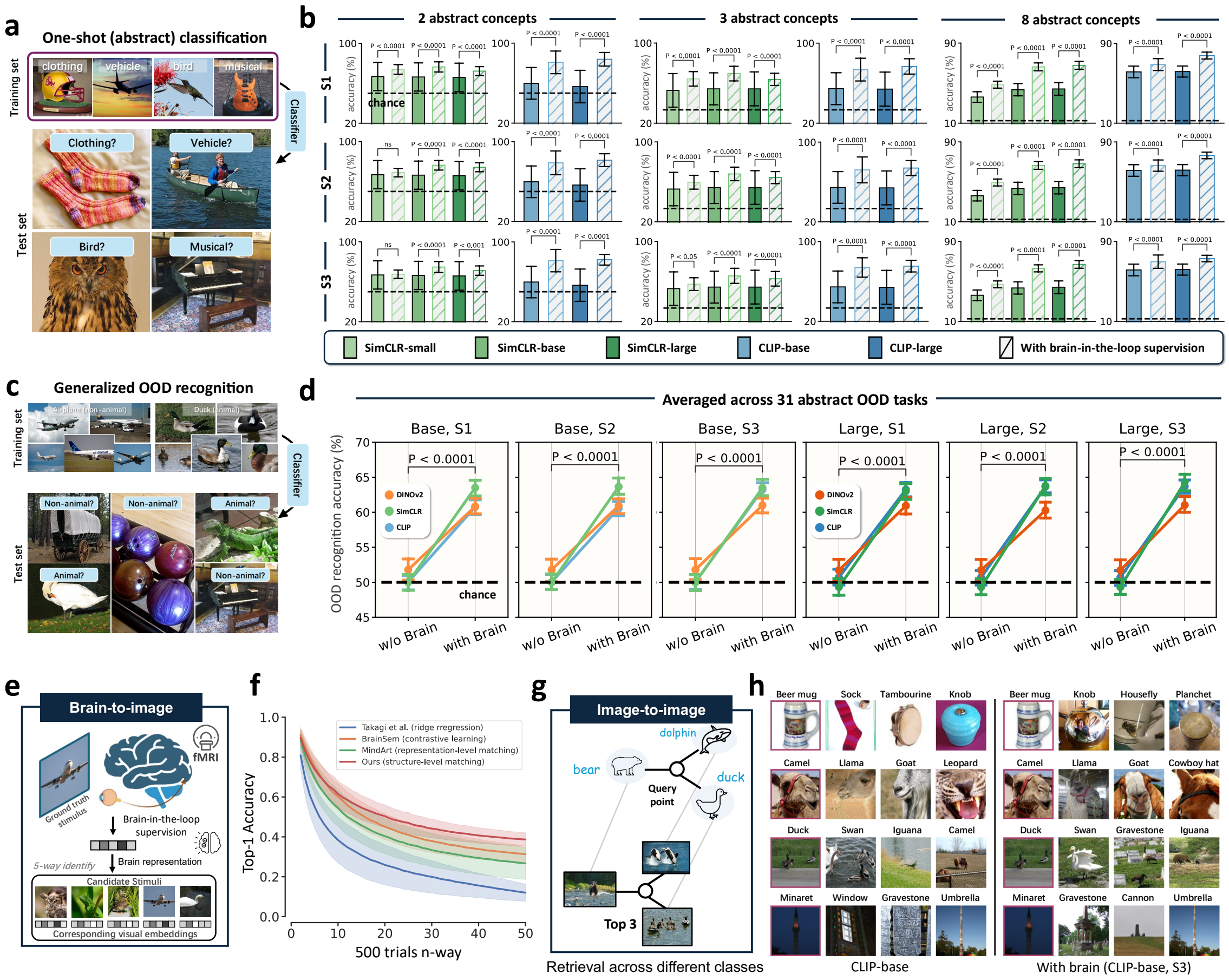

**Fig. 5 | DNNs with human-like conceptual hierarchy obtained enhanced performance on diverse complex tasks. a.** One-shot classification of abstract concepts. The task requires generalizing conceptual knowledge from minimal training examples—for instance, correctly categorizing socks as clothing when provided with only a single exemplar (e.g., an image of a helmet) from the clothing category. **b.** Performance on one-shot classification across three abstraction levels. The brain-in-the-loop supervised DNN obtained significant improvements over diverse DNNs (2 abstract concepts: $P < 0.0001$ for both except SimCLR-large (S3, $P < 0.001$) and SimCLR-small (S2, S3, $P =$ ns), 3 abstract concepts: $P < 0.0001$ for both except SimCLR-small (S3, $P < 0.05$), 8 abstract concepts: $P < 0.0001$ for both, two-tailed paired $t$-test, n = 100 independent runs for each task; see Supplementary for details). See Supplementary Fig. S4a, b for more results. **c.** Generalized OOD recognition tasks. The task assesses the conceptual extrapolation ability of DNNs from minimal category supervision—for example, correctly categorizing bowling as non-animal when given only the knowledge that an airplane is non-animal and a duck is animal. **d.** Performance on generalized OOD recognition across 31 abstract concepts (Supplementary Tab. 2). Brain-in-the-loop supervision significantly improved OOD robustness for different DNNs, yielding performance gains of 11.5 ± 0.2% (CLIP-base), 13.6 ± 0.9% (SimCLR-base), 9.1 ± 0.7% (DINOv2-base), 12.8 ± 0.5% (CLIP-large), 14.4 ± 0.7% (SimCLR-large), and 9.0 ± 0.5% (DINOv2-large) (mean ± s.d. averaged across S1-S3; $P < 0.0001$ for both, two-tailed paired $t$-test, n = 100 independent runs per task; see Supplementary for details). Note that error bars here represent the 99% confidence intervals (CIs). **e.** Brain-to-image retrieval tasks require decoding the semantic content of perceived visual stimuli directly

from brain activity. **f.** Performance of brain-to-image retrieval on 50 stimulus images averaged across S1-S3 (n = 500 trials, see Supplementary). See Supplementary Fig. S6a-c for brain-to-image retrieval cases. **g.** Image-to-image retrieval across different concepts. The task provides a means to qualitatively evaluate the representational structure within the latent space. For example, whether an image of a cup can retrieve semantically related images—such as a teapot—from distinct categories. **h.** Retrieval examples of CLIP-base with (right) and without (left) brain-in-the-loop supervision.

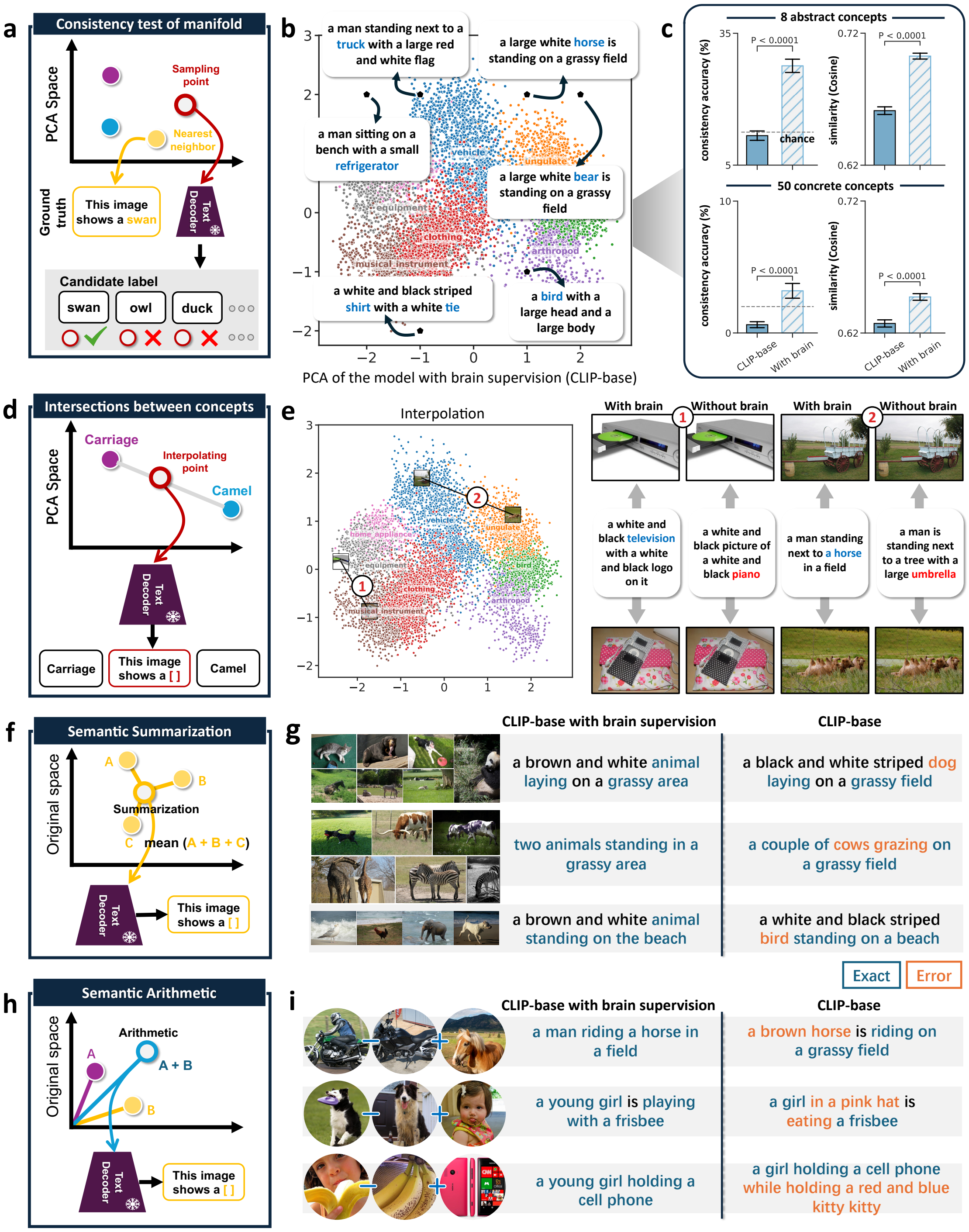

**Fig. 6 | Conceptual manifold with cognitive-level consistency accounts for gains in cognitive generalization. a.** Local semantic consistency test on PCA-based manifold. The task evaluates whether brain-in-the-loop supervision produces a low-dimensional manifold that exhibits meaningful high-level semantic concept boundaries—for instance, vehicle-related concepts can be reconstructed from regions surrounding an airplane sample. **b.** Semantic reconstruction examples of CLIP-base with brain-in-the-loop supervision (S3). We plotted the positions of 8 abstract concepts (see Supplementary). See Supplementary Fig. S7a for reconstruction examples from the CLIP-base without supervision. **c.** Comparison between DNNs with and without brain-in-the-loop supervision on consistency accuracy and semantic

similarity. Brain-in-the-loop supervision (S3) obtained significant improvements over the CLIP-base without supervision ($P < 0.0001$ for both, two-tailed paired $t$-test, n = 100 independent trials, sampling 1,000 2-D data points per trial; see Supplementary). **d.** Semantic reconstruction from intersections between different concepts. The task requires the low-dimensional manifold to support smooth semantic transitions between two concrete concepts. **e.** Semantic reconstruction examples at the intersections between different concepts. **f.** Semantic summarization (operated on original spaces). The task was used to assess whether the model can form a stable conceptual center. **g.** Semantic summarization examples of CLIP-base with (left) and without (right) brain-in-the-loop supervision (S3). **h.** Semantic arithmetic (operated on original spaces). The task assumes that semantic concepts in the latent space can be combined through linear operations. **i.** Semantic arithmetic examples of CLIP-base with (left) and without (right) brain-in-the-loop supervision (S3).

## Methods

## Datasets

**fMRI data.** BOLD signals are obtained from an open dataset DIR[38], which are collected using a Siemens MAGNETOM Verio scanner 3.0-Tesla from three healthy subjects. In image presentation experiments, each subject viewed 1,250 unique images of natural scenes. These image stimuli are drawn from 200 representative object categories of the ImageNet dataset[32], with 150 categories used for training and 50 for testing. Each training image stimulus (8 images per category, totaling 1,200 images) is repeated 5 times, and each test image stimulus (1 image per category) is repeated 24 times, leading to 6,000 and 1,200 image-fMRI instances, respectively. More fMRI data preprocessing details can be found in the original paper. In this study, we focused on voxels from brain visual areas including the lateral occipital complex (LOC), fusiform face area (FFA), and parahippocampal place area (PPA), which are known to be part of the higher visual cortex (HVC)[28].

**Natural images.** The natural images used in this study were drawn from three datasets: ImageNet-21K[32], the THINGS object concept and image dataset[39], and the Common Objects in Context (COCO) dataset[34]. **1)** Corresponding to the BOLD signals of the DIR[38], we collected natural images including 200 object categories from Image-21K for brain-in-the-loop supervision of models, where 150 categories, comprising 206,848 object images, were used as the training set, while the remaining 50 categories, totaling 66,067 object images (275–2115 images per category), constitute the test set. Note that the object categories are the same as those in the DIR, and there is no overlap between the training and test sets. **2)** The images of 1,854 diverse object concepts, collected from the THINGS dataset, were leveraged to perform triplet odd-one-out tasks. **3)** The COCO dataset, a large-scale image dataset including detailed annotations of the images such as image captions, was used to train the text decoder in our manifold analysis experiments.

### Brain-in-the-loop supervised learning

To integrate the human conceptual knowledge into LMs, here we propose a new brain-in-the-loop supervised learning framework to bridge the gap of conceptual structures between DNNs and the brain. Our framework leverages non-invasive neural responses (fMRI) to reshape DNNs, aligning their latent representation structure of concepts with that of the brain (Fig. 2a, b, and Supplementary Fig. S8).

**Problem formulation and notation.** We will denote the model (image) embedding and the fMRI embedding as $x$ and $y$, respectively. Together, $(x, y)$ form a positive or negative pair based on whether their semantic labels match. Given an arbitrary set of image embeddings $\mathbb{X} = \{x_i\}_{i=1}^{n}$ and neural patterns $\mathbb{Y} = \{y_i\}_{i=1}^{m}$, our goal is to find a mapping $f_\theta^*$ such that $f_\theta^* = \underset{f_\theta}{argmin} f_\theta(\mathbb{X}; \mathbb{Y}, d)$, where $d$ represents a structural distance measure that assigns smaller values to more similar structures.

By representing image embeddings and fMRI signals as entities (nodes) and modeling them as graphs, our target can be naturally formulated as an instantiation of graph matching (GM), a technique that aims to find meaningful node correspondences between graphs by considering a structural similarity metric[30]. Our learning procedure is intuitive, comprising two key steps executed alternately: 1) constructing a set of putative matches (correspondences) $\boldsymbol{D} = \{(x_i, y_i)\}_{i=1}^{N}$ from $(\mathbb{X}, \mathbb{Y})$; 2) optimizing the structural similarity between corresponding entities (Supplementary Fig. S8).

**Correspondence construction via optimal transport (OT).** OT provides a means to measure distances between probability distributions over the domain. It is well known that, within the basic form of OT, probability distributions are interpreted as piles of sand. In this context, the distance between two piles of sand can naturally be formulated as the amount of work required to transform one pile into the other, also known as the Earth Mover's Distance (EMD) or Wasserstein Distance (WD)[40]. From a maximum entropy perspective, Cuturi[41] proposed an entropy-regularized OT formulation, allowing the optimal transport plan can be efficiently solved on GPU by Sinkhorn's matrix scaling algorithm. Below, we introduce how to incorporate the OT tools into our brain-in-the-loop supervised learning framework to construct a set of putative matches.

Formally, suppose $\mathbb{X}$ and $\mathbb{Y}$ be sample points in two complete separable metric spaces respectively, and their discrete probability measures can be formulated as $\boldsymbol{\mu} = \sum_{i=1}^{n} p_i\, \delta_{x_i}$ and $\boldsymbol{\nu} = \sum_{i=1}^{m} q_i\, \delta_{y_i}$, where $\delta_{x_i}$ denotes the Dirac function centered on $x_i$, $[p_i]_{1:n} \in \Delta_n$ and $[q_i]_{1:m} \in \Delta_m$ belong to the n- and m-dimensional simplex. Based on this definition, we adopt OT tools[40] to produce a set of putative correspondences by seeking the least costly transport plan $\boldsymbol{\Gamma}_W$, which can be compactly written as:

$$\mathcal{D}_W(\boldsymbol{\mu}, \boldsymbol{\nu}) := \inf_{\gamma \in \Pi} \mathbb{E}_{(\mathbb{X},\mathbb{Y})\sim\gamma}[c(\mathbb{X}, \mathbb{Y})] = \min_{\boldsymbol{\Gamma}_W \in \Pi} \langle \boldsymbol{\Gamma}_W, \boldsymbol{C} \rangle,$$

where $\Pi(\boldsymbol{\mu}, \boldsymbol{\nu}) := \{\boldsymbol{\Gamma}_W \in \mathbb{R}_+^{n\times m} \mid \boldsymbol{\Gamma}_W \mathbb{1}_m = \boldsymbol{\mu}, \boldsymbol{\Gamma}_W{}^{\mathrm{T}} \mathbb{1}_n = \boldsymbol{\nu}\}$, $\boldsymbol{C} \in \mathbb{R}_+^{n\times m}$ is a cost matrix between $\mathbb{X}$ and $\mathbb{Y}$ (calculated by the cosine distance function $c$), and $\langle\cdot,\cdot\rangle$ denotes Frobenius inner product.

**Local structure definition based on re-parametrization techniques.** Suppose we have obtained a set of putative correspondences $\boldsymbol{D} = \{(x_i, y_i)\}_{i=1}^{N}$ between image embeddings $\mathbb{X}$ and neural patterns $\mathbb{Y}$ using the Sinkhorn algorithm, which can be regarded as a node matching procedure. Next, our goal is to maximize the structural similarity between $\mathbb{X}$ and $\mathbb{Y}$ by preserving local semantic structure consensus. Technically, we can leverage argmax function to define neighborhood structure under the cosine distance. However, such a nearest neighbor strategy is not differentiable. Thus, here we employ re-parametrization techniques (i.e., hard Gumbel-Softmax trick[42]) to fulfil differentiable neighborhood structure construction. Formally, we compute the representation similarity score vector $s_i \in \mathbb{R}^{N-1}$ for $x_i$ via Gumbel-Softmax:

$$[s_i]_j = \frac{\exp(x_i \cdot x_j + \gamma_i)}{\sum_{j\neq i} \exp(x_i \cdot x_j + \gamma_i)},$$

where $\{\gamma_i\}_{i=1}^{N-1} \sim$ i.i.d. Gumbel (0, 1). By using the straight through trick, the similarity score vector $s_i$ can be converted into one-hot assignment vector $s_i^*$:

$$s_i^* = \boldsymbol{F}_{\mathrm{argmax}}(s_i) + s_i - \boldsymbol{F}_{\mathrm{sg}}(s_i),$$

where $\boldsymbol{F}_{\mathrm{argmax}}(\cdot)$ returns the hard one-hot version of $s_i$, and $\boldsymbol{F}_{\mathrm{sg}}(\cdot)$ denotes the stop gradient operator. Similarly, we obtain the score vector $z_i \in \mathbb{R}^{N-1}$ for $y_i$ using the same operation.

**Representational structure alignment.** With the local structure defined, structure similarity $\mathcal{L}$ can be quantified from two aspects: 1) the number of correct putative correspondences within local neighborhoods, and 2) the topological similarity among these matching nodes (Supplementary Fig. S8), which can be written as:

$$\mathcal{L}(\mathbb{X}, \mathbb{Y}; \Theta) = \sum_{i=1}^{N} |\, g_k(x_i, y_i; \boldsymbol{D})| + \sum_{i=1}^{N} \mathcal{D}_{gw}\,(x_i, y_i; g_k)$$

where $g_k$ returns the *k* nearest neighbor for each node ($|\cdot|$ returns the cardinality of correct putative correspondences), $\mathcal{D}_{gw}$ denotes a topological metric function. Due to the non-differentiability of the $|\cdot|$ operation, we achieved a similar effect by aligning the sampled distributions between $\mathbb{X}$ and $\mathbb{Y}$. Accordingly, our final cost function can be optimized through the update of model parameters $\Theta$ during end-to-end training. Moreover, for this work, we consider the Gromov-Wasserstein (GW) distance[24] as the topological metric function, which can be used to evaluate the structural similarity between different metric spaces without directly calculating distances between entities across domains. Below, we present how to use GW distance under OT theory to align representational structures between the brain and the model. Note that our optimization process does not rely on task-specific loss functions such as classification.

**Gromov-Wasserstein distance.** The Gromov–Wasserstein (GW) distance generalizes classic OT by evaluating metric spaces directly instead of points across the domains. That is, this approach can be leveraged to operate on distances between point pairs within each domain, as well as measuring how these distances compare to those in the counterpart domain. Therefore, we can use the GW framework to align structured data by preserving the intrinsic topological information.

Let $\boldsymbol{X}_i = \{\hat{x}_i\}_{i=1}^{K_1}$ and $\boldsymbol{Y}_i = \{\hat{y}_i\}_{i=1}^{K_2}$ be the local neighborhoods of $x_i$ and $y_i$, respectively. Then, under GW framework, we can define two measured self-similarity matrices $(\boldsymbol{C}_x, \boldsymbol{p}) \in \mathbb{R}_+^{K_1 \times K_1} \times \Sigma_{K_1}$ and $(\boldsymbol{C}_y, \boldsymbol{q}) \in \mathbb{R}_+^{K_2 \times K_2} \times \Sigma_{K_2}$ for $\boldsymbol{X}_i$ and $\boldsymbol{Y}_i$ using cosine distance, respectively. Since no prior information is available, we set $\boldsymbol{p} = \frac{1}{K_1}\mathbb{1}_{K_1}$ and $\boldsymbol{q} = \frac{1}{K_2}\mathbb{1}_{K_2}$ as uniform distributions. The Gromov–Wasserstein discrepancy between two measured matrices $(\boldsymbol{C}_x, \boldsymbol{p})$ and $(\boldsymbol{C}_y, \boldsymbol{q})$ can be formulated as follows:

$$\mathcal{D}_{gw}(\boldsymbol{C}_x, \boldsymbol{C}_y, \boldsymbol{p}, \boldsymbol{q}) = min_{\boldsymbol{\Gamma}_{GW} \in \Pi(\boldsymbol{p},\boldsymbol{q})} \sum_{i,j,k,l} L\left([\boldsymbol{C}_x]_{i,k}, [\boldsymbol{C}_y]_{j,l}\right)[\boldsymbol{\Gamma}_{GW}]_{i,j}[\boldsymbol{\Gamma}_{GW}]_{k,l},$$

where $\Pi(\boldsymbol{p}, \boldsymbol{q})$ denotes all the admissible couplings between $\boldsymbol{p}$ and $\boldsymbol{q}$, $\boldsymbol{\Gamma}_{GW}$ is a coupling (transport plan) between the two spaces, and $L: \mathbb{R} \times \mathbb{R} \rightarrow \mathbb{R}$ is a loss function to evaluate the misfit between two pairs of entities $(\hat{x}_i,\ \hat{x}_k)$ and $(\hat{y}_j,\ \hat{y}_l)$. In the case $L(a, b) = L_2(a, b) = \frac{1}{2}(a - b)^2$, $GW^{1/2}$ defines a distance on the space of metric measure spaces. Intuitively, the loss term $L([\boldsymbol{C}_x]_{i,k}, [\boldsymbol{C}_y]_{j,l})$ can also be understood as the cost of matching graph edges $(\hat{x}_i,\ \hat{x}_k)$ and $(\hat{y}_j,\ \hat{y}_l)$ across two domains.

Taken together, with the help of the entropy-regularized OT formulation and the re-parametrization technique, all operations in the brain-in-the-loop supervised learning framework are differentiable, enabling us to maximize the structural similarity between $\mathbb{X}$ and $\mathbb{Y}$ through backpropagation to update the latent embeddings during training.

**Architecture and optimization**

**Model architecture**. Technically, like most representation learning practices, our first component employs two nonlinear encoders (here, a ViT[43] and a multilayer perceptron but can take any other form) to map brain signals (fMRI) and image embeddings to new latent spaces (Fig. 2a). Next, a shared multiplex attention GNN[30,44,45] is leveraged to predict an assignment matrix with minimal Wasserstein distance[40], resulting in initial correspondences between brain and model representations. The representations in latent spaces are considered as graph nodes (entities). To reshape embedding spaces of DNNs, we leverage a differentiable sampler (based on re-parametrization technique Gumbel Softmax[42]) with the similarity scores between entities to construct local structures for each putative correspondence. It is worth noting that the procedure of structure (edge) construction is "*dynamic*" since node representations are evolving through the update of neural network parameters during training. See Supplementary for the architecture configuration and hyperparameter settings.

By using relaxation techniques[41,42], all calculation and optimization processes in the pipeline are differentiable, enabling gradients to be backpropagated from the matching layers to the encoders. We conduct brain-in-the-loop supervision on the DIR dataset. The stimuli (150 training categories and 50 test categories) in DIR dataset are identical to those used in the GOD dataset[28] but contain a larger number of evoked fMRI recordings (6,000 training samples and 1,200 test samples). See Supplementary for details of training procedures.

**Loss function.** Given an image-fMRI dataset $(\mathbb{X}, \mathbb{Y})$, we first minimized the negative log-likelihood of the transport plan $\boldsymbol{\Gamma}_W$, and a cross-entropy item to seek a set of putative correspondences $\boldsymbol{D}$:

$$\mathcal{L}_W = -\sum_{(i,j)\in \boldsymbol{V}} log[\boldsymbol{\Gamma}_W]_{i,j} - \lambda_1 \sum_{(i,j)\in \boldsymbol{V}} log[softmax(\boldsymbol{C})]_{i,j}$$

where $\boldsymbol{V}$ contains ground-truth matching labels, $\lambda_1 = 0.1$ is used to balance the two terms, and $\boldsymbol{C}$ is the cosine similarity matrix derived from the outputs of the image and fMRI encoders. The transport plan $\boldsymbol{\Gamma}_W = \text{diag}(\mathbf{a})\mathbf{K}\text{diag}(\mathbf{b})$ was obtained through the solution of an entropy-regularized OT problem $min\langle \boldsymbol{\Gamma}_W, \mathbf{1} - \boldsymbol{C}\rangle - \varepsilon H(\boldsymbol{\Gamma}_W)$, where $H(\boldsymbol{\Gamma}_W) = -\sum_{(i,j)}[\boldsymbol{\Gamma}_W]_{i,j} \, log[\boldsymbol{\Gamma}_W]_{i,j}$ is an entropy constraint term (trade-off parameter $\varepsilon = 1$ ), $\boldsymbol{K} = e^{-(1/\varepsilon)\boldsymbol{C}} \in \mathbb{R}_+^{n\times m}$ is a Gibbs kernel, $\boldsymbol{a} \in \mathbb{R}_+^n$ and $\boldsymbol{b} \in \mathbb{R}_+^m$ can be computed by Sinkhorn iterations[41].

Let $s_i^*$ and $z_i$ be the hard and soft one-hot vectors for $x_i$ and $y_i$, respectively, computed using Gumbel-Softmax. Then, we minimized the cross-entropy loss between $s_i^*$ and $z_i$, and GW distance between dynamically constructed local structures, which can be written as:

$$\mathcal{L}_{GW} = -\sum_{j=1}^{N-1} [s_i^*] \, log[z_i]_j + \frac{\lambda_2}{W}\sum_{V} L\left([\boldsymbol{C}_x]_{i,k}, [\boldsymbol{C}_y]_{j,l}\right) [\boldsymbol{\Gamma}_{GW}]_{i,j}[\boldsymbol{\Gamma}_{GW}]_{k,l}.$$

Similarly, the transport plan $\boldsymbol{\Gamma}_{GW}$ was derived by solving an entropy-regularized optimization problem using the Sinkhorn algorithm. In the second item, we only penalize the false correspondences $V$, where $W \leq min(K_1, K_2)$ denotes the number of true correspondences within constructed local structures (trade-off parameter $\lambda_2$ = 100). Together, $\mathcal{L}_W$ and $\mathcal{L}_{GW}$ constituted the final loss functions of the brain-in-the-loop supervision framework across the two iterative steps.

## Statistical analysis

In this study, we applied the two-tailed paired *t*-test for statistical analysis. For Fig. 1a (top, without brain-in-the-loop supervision), the *P* values are $3.56 \times 10^{-87}$/$1.55 \times 10^{-30}$ (SimCLR-small vs SimCLR-large/SimCLR-base vs SimCLR-large), $1.78 \times 10^{-76}$/$6.55 \times 10^{-40}$ (DINOv2-small vs DINOv2-large/DINOv2-base vs DINOv2-large) and $8.72 \times 10^{-65}$ (CLIP-base vs CLIP-large) respectively, both with n = 100. In Fig. 1a (bottom, with brain-in-the-loop supervision), the *P* values are $1.69 \times 10^{-102}$/$1.07 \times 10^{-43}$ (SimCLR-small vs SimCLR-large/SimCLR-base vs SimCLR-large), $8.16 \times 10^{-47}$/$1.85 \times 10^{-11}$ (DINOv2-small vs DINOv2-large/DINOv2-base vs DINOv2-large) and $2.38 \times 10^{-40}$ (CLIP-base vs CLIP-large) respectively, both with n = 100. In Fig. 1b, the *P* values for the models without brain-in-the-loop supervision are $9.54 \times 10^{-38}$/$2.84 \times 10^{-5}$ (SimCLR-small vs SimCLR-large/SimCLR-base vs SimCLR-large), $6.66 \times 10^{-4}$/$5.75 \times 10^{-3}$ (DINOv2-small vs DINOv2-large/DINOv2-base vs DINOv2-large) and 0.302 (CLIP-base vs CLIP-large) respectively; the *P* values for the models with brain-in-the-loop supervision (S3) are $7.96 \times 10^{-60}$/$9.28 \times 10^{-12}$ (SimCLR-small vs SimCLR-large/SimCLR-base vs SimCLR-large), $2.61 \times 10^{-1}$/$1.51 \times 10^{-4}$ (DINOv2-small vs DINOv2-large/DINOv2-base vs DINOv2-large) and $1.91 \times 10^{-5}$ (CLIP-base vs CLIP-large) respectively, both with n = 100. In Fig. 3a, the *P* values are $1.44 \times 10^{-75}$/$4.24 \times 10^{-57}$, $4.32 \times 10^{-89}$/$6.16 \times 10^{-90}$ and $3.07 \times 10^{-93}$/$1.56 \times 10^{-90}$ (CLIP-base/CLIP-large vs CLIP-base with brain-in-the-loop supervision) for 50 concrete concepts, 8 abstract concepts, and 3 abstract concepts, respectively (n = 50 independent runs). In Fig. 4d, the *P* values are $1.78 \times 10^{-82}$/$2.95 \times 10^{-163}$, $2.64 \times 10^{-83}$/$4.56 \times 10^{-271}$ and $3.30 \times 10^{-83}$/$1.63 \times 10^{-277}$ (matrix-wise/row-wise) for S1-S3, respectively (n = 50). In Fig. 4f, the *P* values (n = 100) are $7.56 \times 10^{-97}$/$3.70 \times 10^{-95}$/$1.01 \times 10^{-121}$ (CLIP-large/SimCLR-large/DINOv2-large vs CLIP-base with brain-in-the-loop supervision), respectively. See Supplementary Tab. 3 for the *P* values in Fig. 5b and Supplementary Fig. S4. In Fig. 5d, the *P* values are $3.22 \times 10^{-74}$/$1.17 \times 10^{-94}$/$8.09 \times 10^{-34}$/$2.33 \times 10^{-111}$/$1.49 \times 10^{-90}$/$1.45 \times 10^{-32}$ (S1), $5.99 \times 10^{-73}$/$5.09 \times 10^{-95}$/$1.84 \times 10^{-34}$/$1.29 \times 10^{-112}$/$3.79 \times 10^{-101}$/$1.29 \times 10^{-27}$ (S2), and (≈) 0/$2.71 \times 10^{-97}$/$5.15 \times 10^{-35}$/$1.11 \times 10^{-121}$/$5.69 \times$

$10^{-110}$/4.74 × $10^{-33}$ (S3) for the models (CLIP-base/SimCLR-base/DINOv2-base/CLIP-large/SimCLR-large/DINOv2-large) with and without brain-in-the-loop supervision (n = 3100). In Fig. 6c, the *P* values are 6.72 × $10^{-94}$/5.32 × $10^{-62}$ and 9.58 × $10^{-108}$/4.70 × $10^{-83}$ (8 abstract concepts/50 concrete concepts) for consistency accuracy and semantic similarity, respectively (n = 100). Data visualizations were accomplished using seaborn, a Python data visualization library, and statannotations, which is used to add statistical significance annotations on seaborn plots.

## Data availability

All data used in this study are publicly available. 1) DIR dataset: https://openneuro.org/datasets/ds001506/versions/1.3.1. We used the officially preprocessed fMRI data (v2 data) from https://figshare.com/articles/dataset/Deep_Image_Reconstruction/7033577; 2) THING database: https://osf.io/f5rn6/; 3) Microsoft COCO 2017 dataset: https://cocodataset.org/; 4) ImageNet-21K dataset: https://image-net.org/download. Download the images of a single synset: https://image-net.org/data/winter21_whole/[synsetid].tar (replace [synsetid] with the actual synset ID). We used Natural Language Toolkit (NLTK) in python to acquire WordNet.

## Author contributions

J.C. proposed the brain-in-the-loop supervised learning architecture, analyzed the data. wrote the paper. Y.Q. conceived the project, designed the experiments, wrote the paper. Y.W. supervised the project. G.P. supervised the project and contributed to the manuscript writing.

## Competing interests

The authors declare no competing interests.

## Acknowledgments

This work was supported in part by the Science and Technology Innovation (STI) 2030 Major Projects No. 2021ZD0200400 (G.P.), Zhejiang Provincial Natural Science Foundation of China No. LR24F020002 (Y.Q.), Natural Science Foundation of China (NSFC) No. 624B2127 (J.C.), and the Fundamental Research Funds for the Central Universities No. 2024BSSXM11 (J.C.). The funders had no role in study design, data collection and analysis, decision to publish or preparation of the manuscript.

# Supplementary Information

## Supplementary Figures

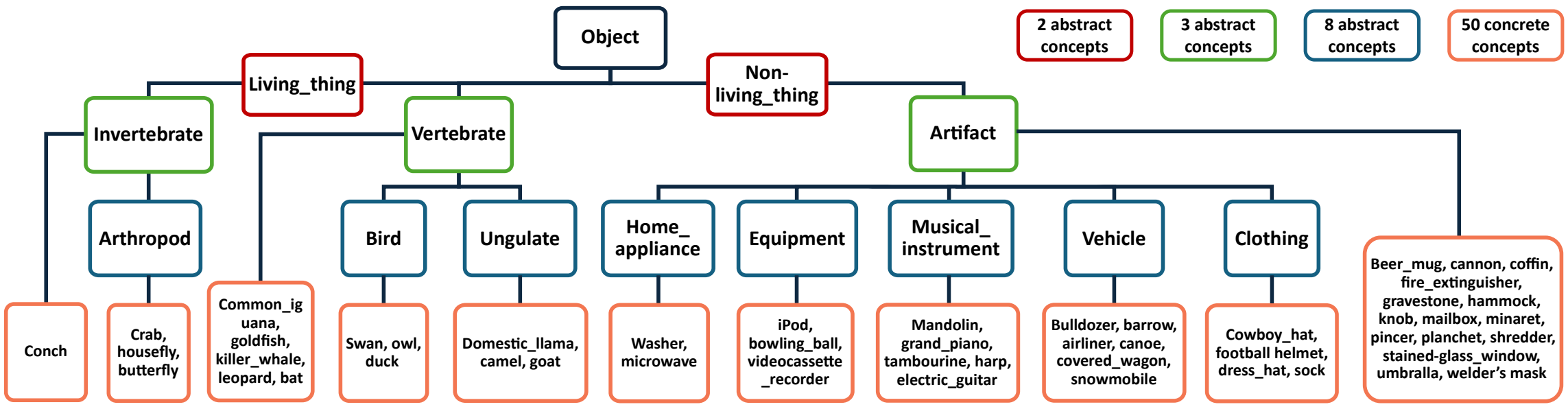

**Supplementary Fig. S1 | Concepts across four abstraction levels.** These abstract concepts are derived from hypernyms (superclasses) of WordNet. Note that image annotations in ImageNet are structured according to the WordNet hierarchy. See Supplementary Tab. 4 for other combinations of abstract concepts used in this study.

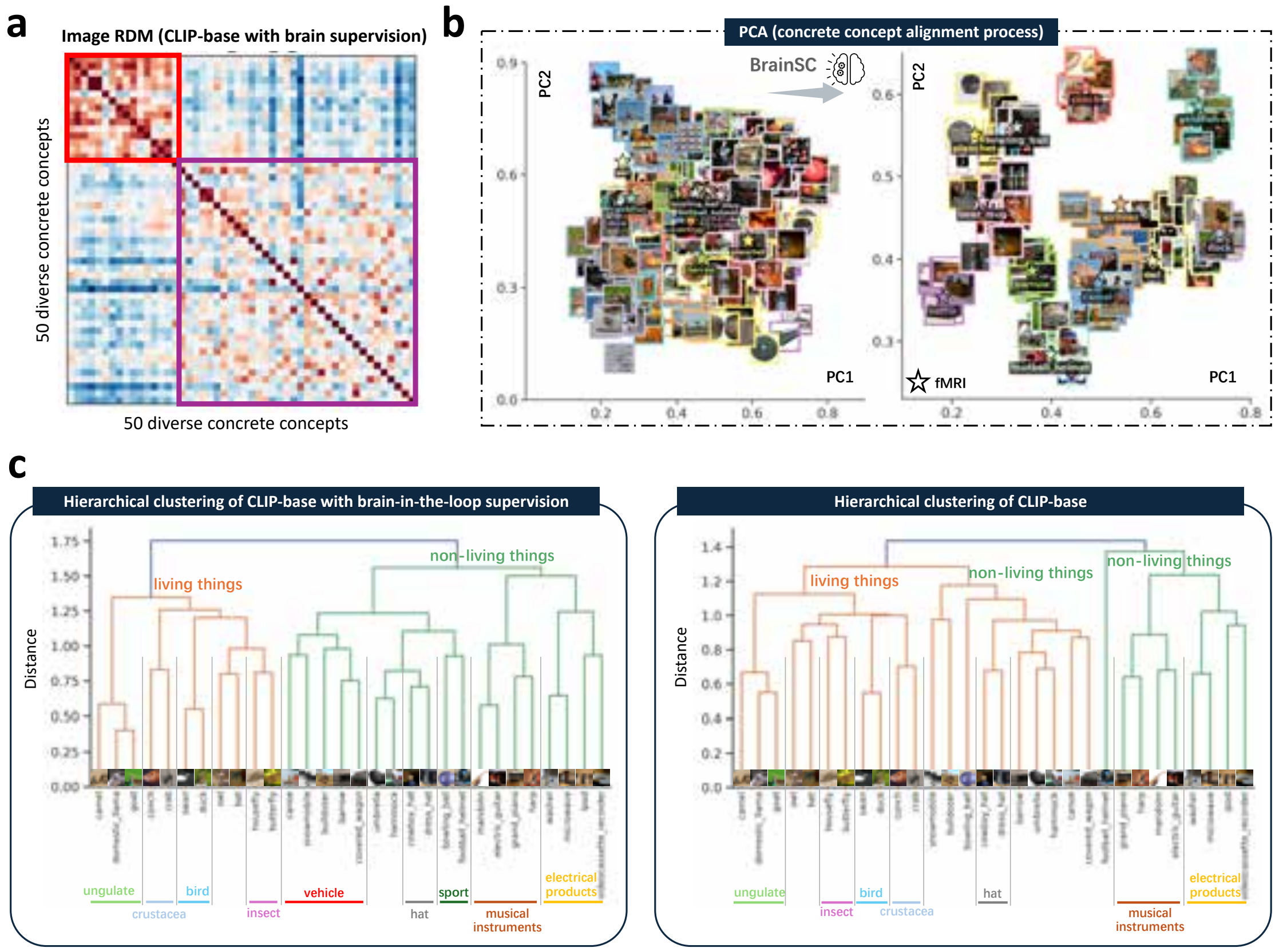

**Supplementary Fig. S2 | Investigating how brain-in-the-loop supervision shapes the conceptual representational structure of DNNs. a,** Representational dissimilarity matrix (RDM) for CLIP-base with brain-in-the-loop supervision (averaged across S1-S3) RDM was computed by using ground truth stimulus images (n = 50). Red square denotes living objects, and purple square is non-living objects. **b,** Comparison of representational spaces before and after brain-in-the-loop supervision (S3). The position of each dot in the visualization is determined by the first two principal components. Visualization shows that the representational structure became more similar to fMRI after brain-in-the-loop supervision. Stars indicate fMRI embeddings, with colors corresponding to ground-truth labels. **c,** Hierarchical clustering of CLIP-base with and without brain-in-the-loop supervision. Hierarchical clustering was performed via the average-linkage hierarchical clustering algorithm (implemented in SciPy, with default parameters) on concept embedding centroids (275–2,115 images per category).

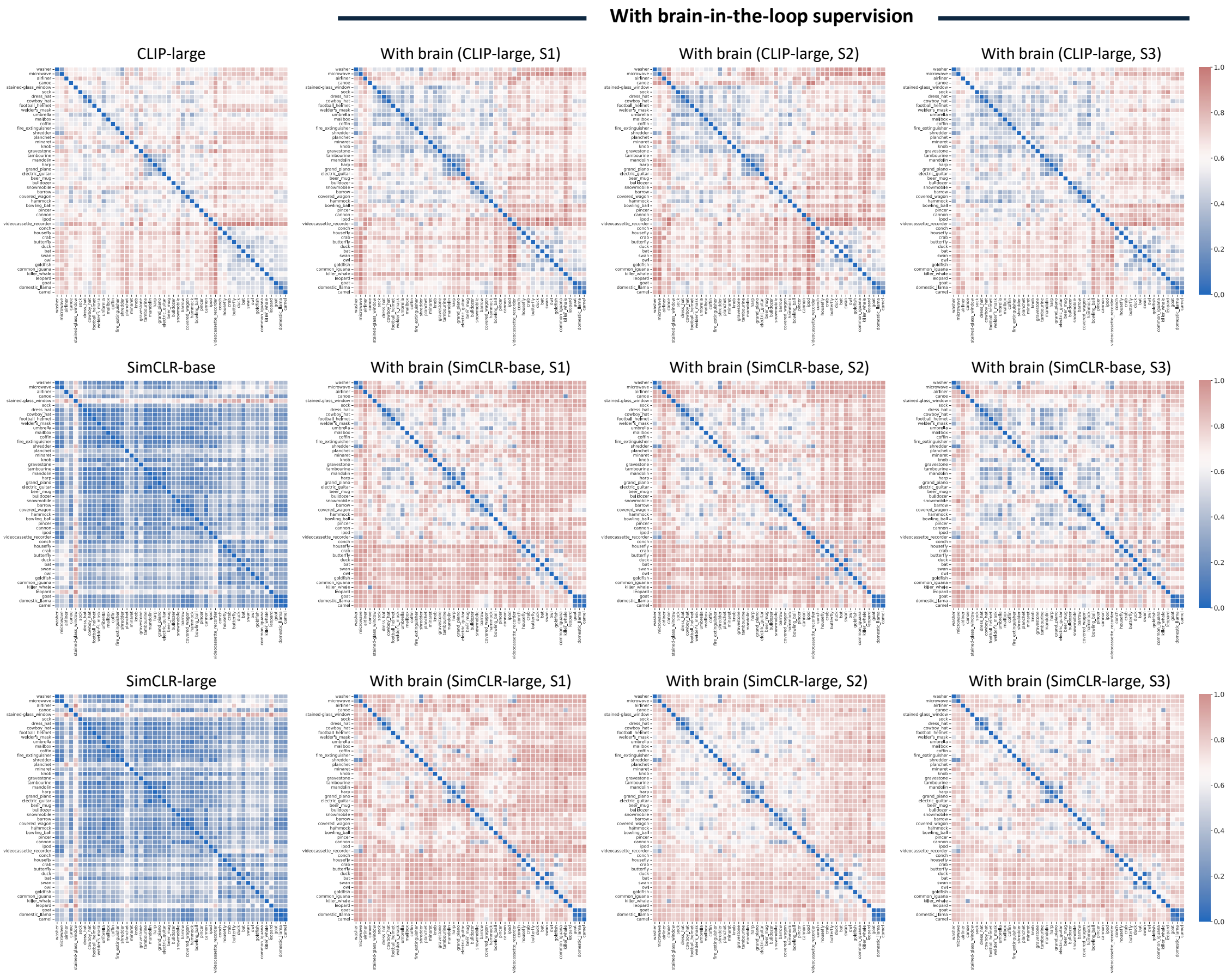

**Supplementary Fig. S3 | Concept similarity matrices (CSMs) of different DNN architectures across varying parameter scales.** To build CSMs for DNNs, we computed the pairwise cosine distances between concept embedding centroids, which were obtained by averaging the image representations within each category (n = 66,067; 275–2,115 images per category). The rows and columns of the DNN CSMs were plotted in the same order as WordNet' s CSM to facilitate comparison (Fig. 4a).

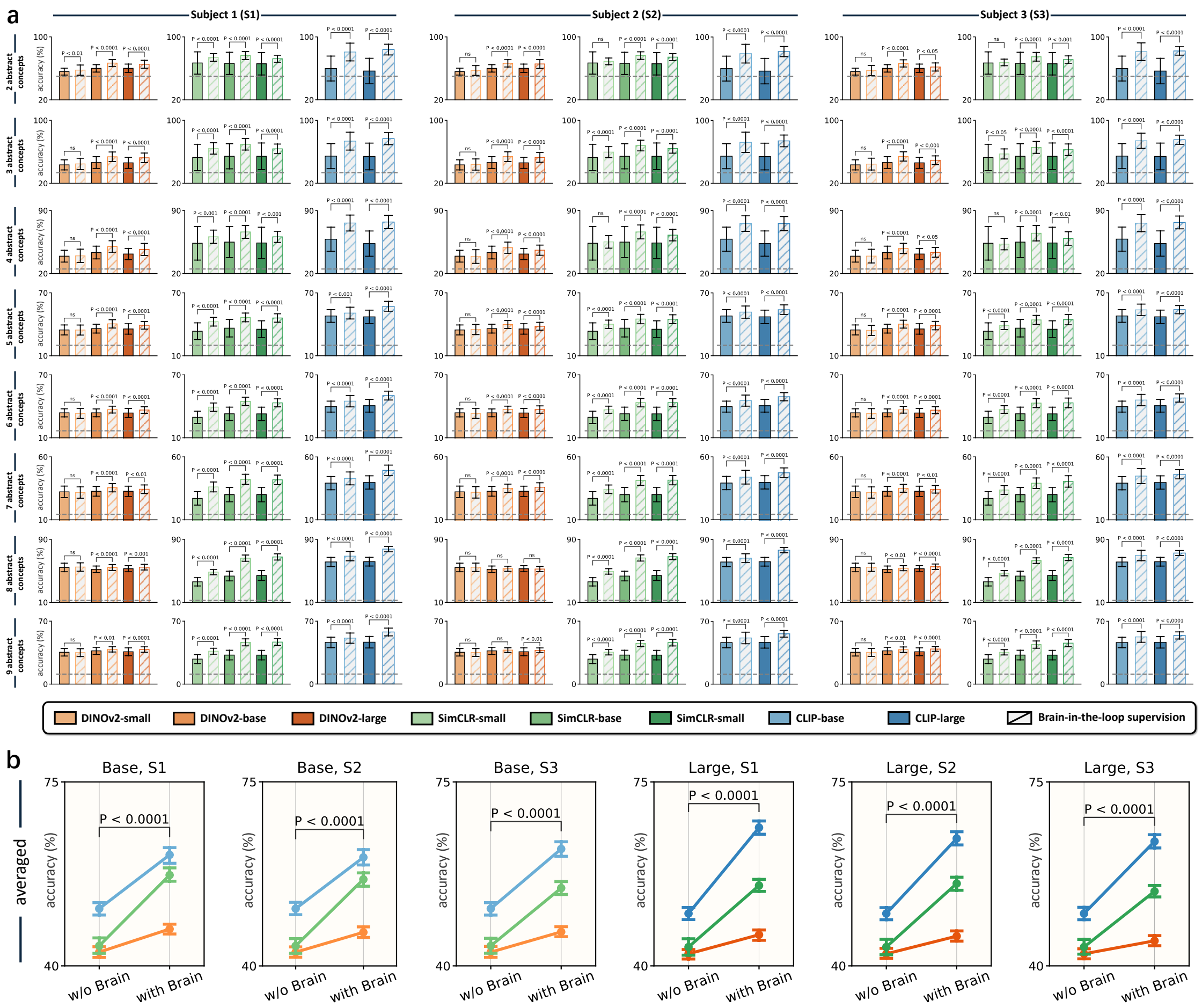

**Supplementary Fig. S4 | Performance on one-shot abstract classification. a,** One-shot accuracy across nine abstraction levels (Supplementary Tab. 4) on different DNNs. For each task, we run 100 independent trials (see Supplementary Text for training details), where each trial includes n = [66,037 – the number of abstract concepts] test samples. **b,** Brain-in-the-loop supervision significantly improved one-shot learning robustness across nine abstraction levels for different DNNs, yielding performance gains of 10.4 ± 0.47%, 12.4 ± 0.17%, 4.0 ± 0.05%, 14.8 ± 0.54%, 11.5 ± 0.27%, and 3.2 ± 0.03% (mean ± s.d.; averaged across 3 subjects; $P < 0.0001$ for both, two-tailed paired $t$-test) for CLIP-base, SimCLR-base, DINOv2-base, CLIP-large, SimCLR-large, and DINOv2-large, respectively. Note that error bars here represent 99% confidence intervals (CIs).

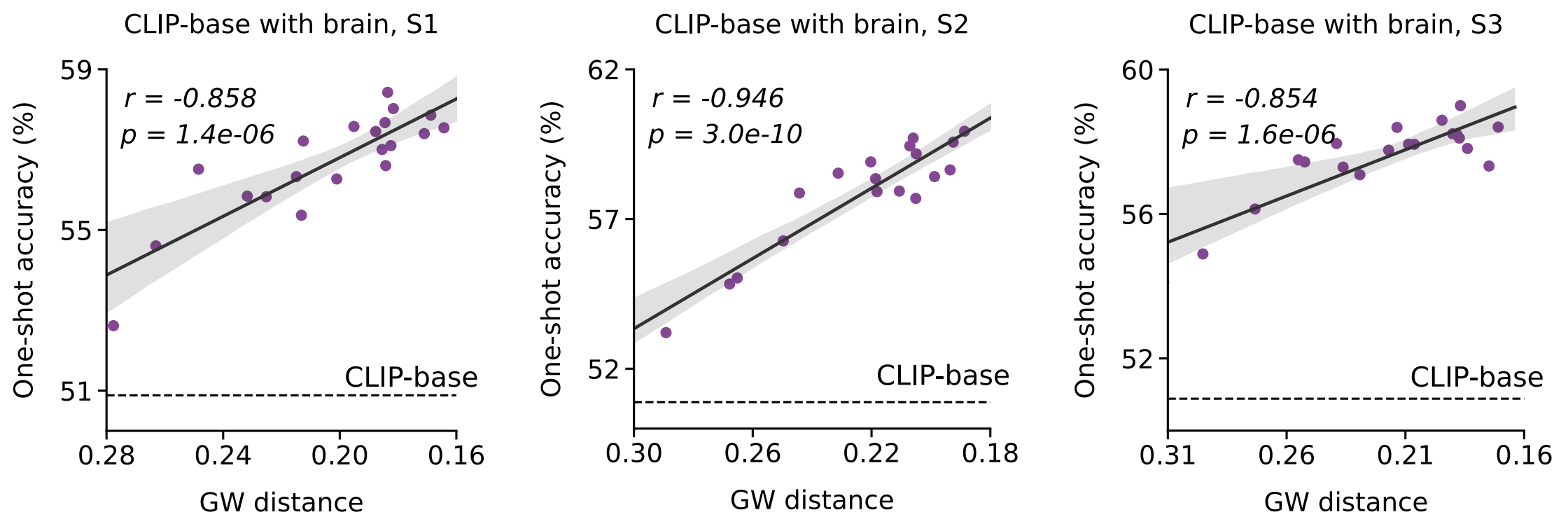

**Supplementary Fig. S5 | One-shot performance was plotted against GW distance between fMRI and their stimulus image embeddings during brain-in-the-loop supervision.** One-shot performance was averaged across eight combinations of abstract concepts (Supplementary Tab. 4; n = 100 independent runs per abstract level). The plot revealed a strong negative correlation, suggesting that better alignment of the representational structure between the brain and the model leads to improved one-shot abstract classification performance**.**

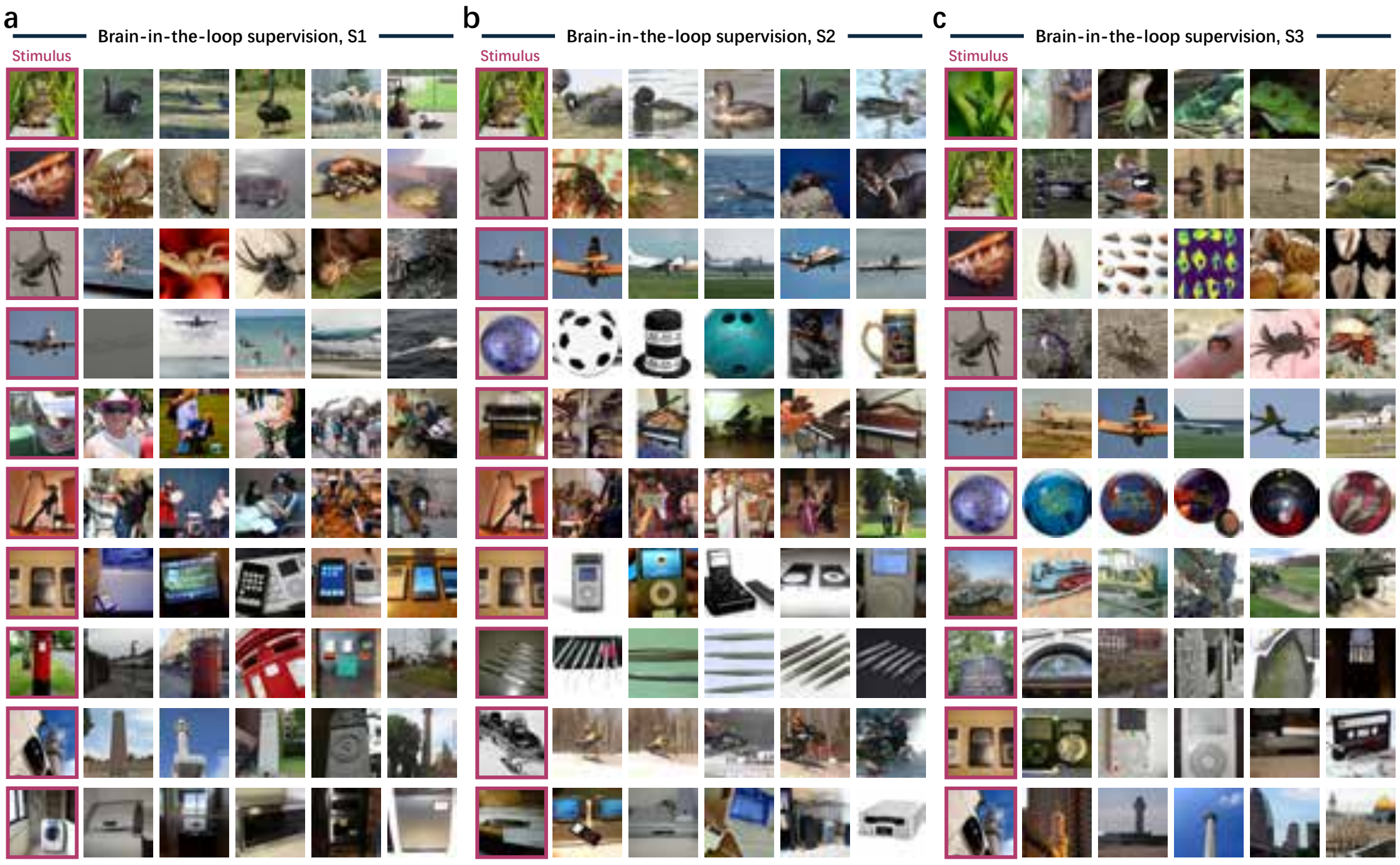

**Supplementary Fig. S6 | Brain-to-image retrieval examples for CLIP-base with brain-in-the-loop supervision. a-c,** Here, brain-to-image retrieval was performed on a subset of WordNet-21K (including 50 concrete concepts; n = 66,067), rather than on 50 real stimuli during fMRI acquisition (Fig. 5f). The image with a purple border represents the ground-truth label (real stimulus).

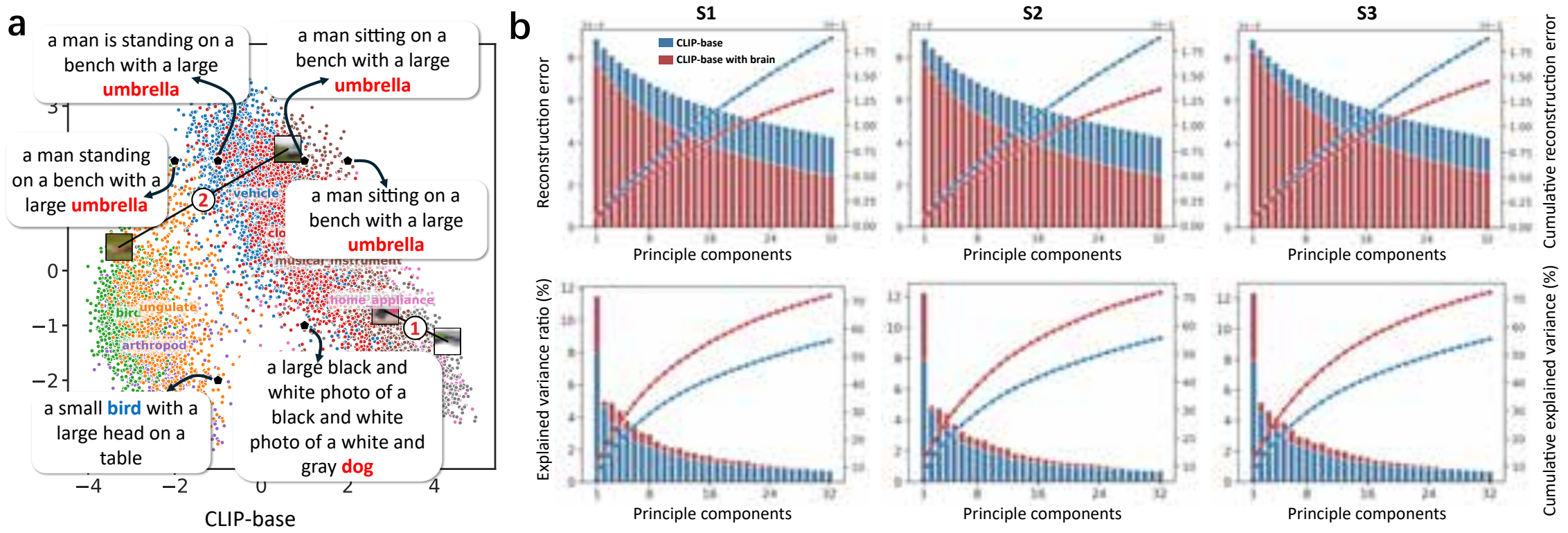

**Supplementary Fig. S7 | Reconstruction error and explained variance on the PCA-based manifold. a,** Semantic reconstruction examples of CLIP-base. Like Fig. 6b, e, the PCA-based manifold here was generated from n = 13,750 test images (50 unseen categories, 275 images per category), and we plotted the positions of 8 abstract concepts. The positions of the concept names were determined based on the centroid of 275 within-class samples. The PCA visualization revealed that, without brain-in-the-loop supervision, representations failed to organize according to high-level semantic concepts. **b,** Comparison of PCA reconstruction error and explained variance ratio (measured on n = 13,750 test images) between CLIP-base with and without brain-in-the-loop supervision.

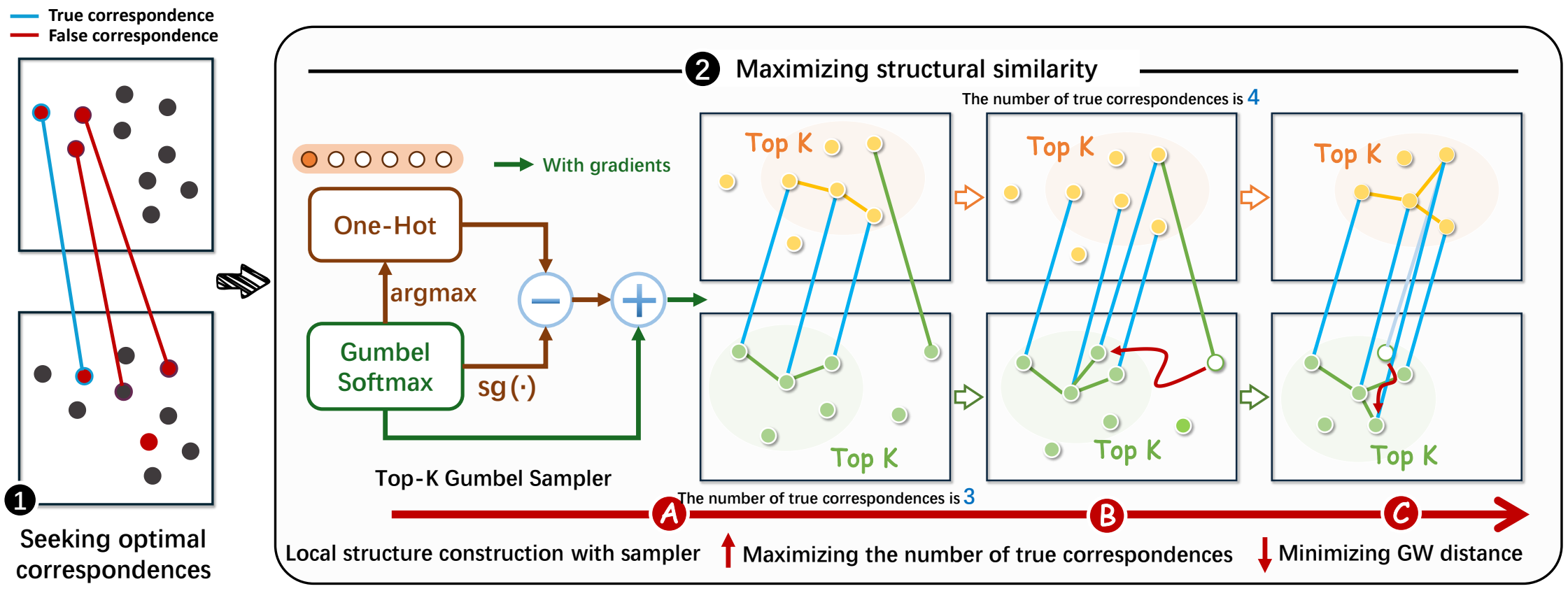

**Supplementary Fig. S8 | An illustration of the key steps involved in the structural similarity maximization procedure.** Suppose we have obtained a set of putative correspondences via OT (Wasserstein distance). We next leverage a differentiable Gumbel sampler to construct local structures for each correspondence (putative matching). By doing so, our optimization goal (maximizing structural similarity) can be formulated as the number of true correspondences within neighborhood structures and the topological similarity of these nodes (measured by GW distance), which encourages the model to maintain local semantic consensus with the conceptual representational structure of humans. These optimization steps are iterative. In other words, after updating the node representations, our brain-in-the-loop supervised learning model will reconstruct the putative matching and optimize the conceptual structures between DNNs and humans.

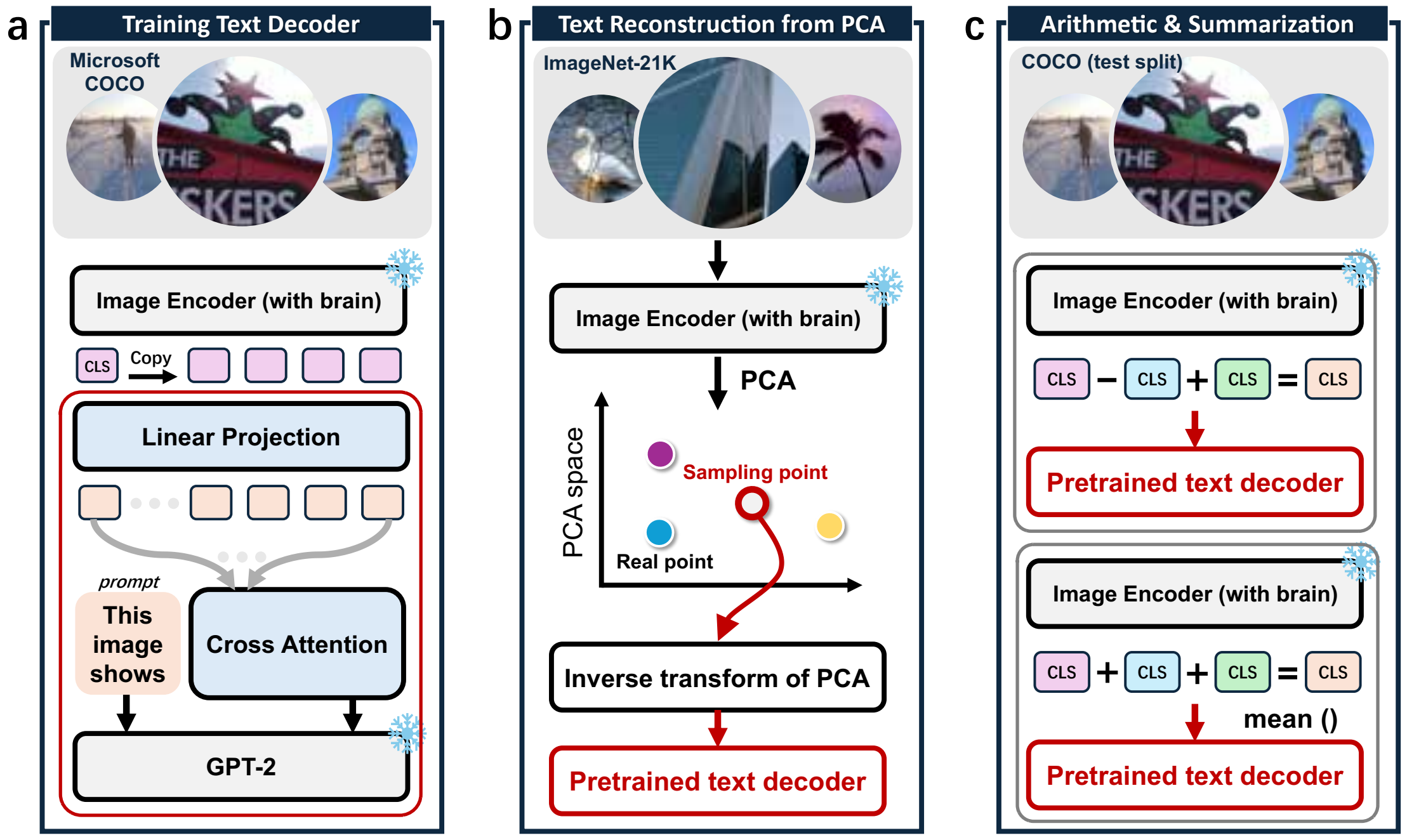

**Supplementary Fig. S9 | Schematic of the main operations involved in the cognitive-level consistency test. a,** Given an image [CLS] embedding, we first employed a learnable linear projection layer to yield a sequence of 768-D embeddings (n = 8). We then used cross-attention layers and a text prompt to bridge these embeddings and GPT-2 model. That is, the GPT-2 model generated text conditioned on a sequence of 768-D embeddings and a prompt "This image shows" (see Supplementary Text for details). **b,** To reconstruct semantics from the manifold, we used PCA to construct a low-dimensional space, which allows us to sample data points from any spatial position and map them back to the original space. Then, these reconstructed embeddings can be fed to the well-trained GPT-2-based decoder for text decoding. **c,** Arithmetic and summarization operations were performed on the original space, rather than on the PCA-based manifold.

# Supplementary Tables

| | | | | |
|---|---|---|---|---|
| 1. washer | 2. microwave | 3. airliner | 4. canoe | 5. stained-glass window |
| 6. sock | 7. dress hat | 8. cowboy hat | 9. football helmet | 10. welder’s mask |
| 11. umbrella | 12. mailbox | 13. coffin | 14. fire extinguisher | 15. shredder |
| 16. planchet | 17. minaret | 18. knob | 19. gravestone | 20. tambourine |
| 21. mandolin | 22. harp | 23. grand piano | 24. electric guitar | 25. beer mug |
| 26. bulldozer | 27. snowmobile | 28. barrow | 29. covered wagon | 30. hammock |
| 31. bowling ball | 32. pincer | 33. cannon | 34. ipod | 35. videocassette recorder |
| 36. conch | 37. housefly | 38. crab | 39. butterfly | 40. duck |
| 41. bat | 42. swan | 43. owl | 44. goldfish | 45. common iguana |
| 46. killer whale | 47. leopard | 48. goat | 49. domestic llama | 50. camel |

**Supplementary Tab. 1 | The concept order used in the WordNet conceptual similarity matrix (CSM).** This ordering was applied to both rows and columns in all CSM visualizations (Fig. 4a-c and Supplementary S3).

| Generalized OOD tasks | Including concrete concepts |
|---|---|
| **insect** | housefly, butterfly |
| **percussion instrument** | grand piano, tambourine |
| **protective covering** | welder's mask, umbrella |
| **home appliance (appliance, durables)** | washer, microwave |
| **box** | mailbox, coffin |
| **hat** | cowboy hat, dress hat |
| **craft** | airliner, canoe |
| **self-propelled vehicle** | bulldozer, snowmobile |
| **chordophone** | mandolin, harp |
| **aquatic bird** | swan, duck |
| **bird** | swan, owl, duck |
| **ungulate (even-toed ungulate)** | domestic llama, camel, goat |
| **headdress** | cowboy hat, football helmet, dress hat |
| **equipment** | ipod, bowling ball, videocassette recorder |
| **arthropod** | crab, housefly, butterfly |
| **invertebrate** | conch, crab, housefly, butterfly |
| **structure** | knob, stained-glass window, gravestone, minaret |
| **wheeled vehicle** | bulldozer, barrow, covered wagon, snowmobile |
| **stringed instrument** | mandolin, grand piano, electric guitar, harp |
| **clothing** | cowboy hat, football helmet, dress hat, sock |
| **musical instrument** | mandolin, grand piano, tambourine, electric guitar, harp |
| **consumer goods (commodity)** | washer, microwave, cowboy hat, football helmet, dress hat, sock |
| **mammal (placental)** | leopard, domestic llama, killer whale, bat, camel, goat |
| **covering** | welder's mask, cowboy hat, football helmet, dress hat, umbrella, sock |
| **vehicle** | bulldozer, barrow, airliner, covered wagon, canoe, snowmobile |
| **container** | mailbox, bulldozer, barrow, beer mug, coffin, covered wagon, snowmobile |
| **device** | fire extinguisher, mandolin, grand piano, tambourine, electric guitar, harp, planchet, shredder |
| **vertebrate (chordate)** | swan, goldfish, leopard, common iguana, owl, domestic llama, killer whale, bat, camel, goat, duck |
| **living thing (animal, organism)** | conch, swan, goldfish, leopard, crab, common iguana, owl, domestic llama, killer whale, housefly, bat, camel, goat, duck, butterfly |
| **instrumentality** | mailbox, fire extinguisher, pincer, bulldozer, mandolin, ipod, grand piano, barrow, tambourine, cannon, electric guitar, beer mug, bowling ball, videocassette recorder, airliner, coffin, covered wagon, harp, planchet, shredder, canoe, snowmobile, hammock |
| **non-living thing (artifact)** | see Supplementary Fig. S1 |

**Supplementary Tab. 2 | Generalized out-of-distribution (OOD) recognition targets.** We used all hypernyms (n = 41) of 50 concrete concepts as OOD recognition targets (Fig. 5d), excluding 9 hypernyms that had duplicate subclasses. Each hypernym included at least two concrete concepts (subclasses).

| Num. (Sub) | CLIP (base/large) | SimCLR (small/base/large) | DINOv2 (small/base/large) |
|---|---|---|---|
| **2 (S1)** | $2.47 \times 10^{-20}/1.03 \times 10^{-28}$ | $1.07 \times 10^{-5}/1.07 \times 10^{-9}/4.81 \times 10^{-5}$ | $8.10 \times 10^{-3}/4.01 \times 10^{-16}/3.33 \times 10^{-11}$ |
| **3 (S1)** | $1.29 \times 10^{-15}/8.28 \times 10^{-20}$ | $5.22 \times 10^{-9}/1.04 \times 10^{-12}/6.20 \times 10^{-8}$ | $2.00 \times 10^{-1}/1.03 \times 10^{-10}/3.74 \times 10^{-9}$ |
| **4 (S1)** | $3.97 \times 10^{-19}/1.09 \times 10^{-26}$ | $4.88 \times 10^{-4}/4.60 \times 10^{-9}/3.76 \times 10^{-4}$ | $7.16 \times 10^{-1}/7.21 \times 10^{-10}/2.48 \times 10^{-7}$ |
| **5 (S1)** | $8.99 \times 10^{-4}/3.73 \times 10^{-24}$ | $8.08 \times 10^{-18}/1.45 \times 10^{-18}/4.27 \times 10^{-20}$ | $8.85 \times 10^{-1}/3.01 \times 10^{-12}/2.58 \times 10^{-7}$ |
| **6 (S1)** | $6.85 \times 10^{-10}/3.93 \times 10^{-24}$ | $2.58 \times 10^{-24}/7.91 \times 10^{-27}/8.17 \times 10^{-27}$ | $3.68 \times 10^{-1}/3.50 \times 10^{-8}/7.30 \times 10^{-7}$ |
| **7 (S1)** | $5.69 \times 10^{-7}/8.33 \times 10^{-24}$ | $2.39 \times 10^{-26}/4.01 \times 10^{-32}/3.47 \times 10^{-31}$ | $3.63 \times 10^{-1}/3.09 \times 10^{-7}/3.31 \times 10^{-3}$ |
| **8 (S1)** | $8.36 \times 10^{-14}/1.36 \times 10^{-45}$ | $2.40 \times 10^{-36}/1.71 \times 10^{-53}/9.58 \times 10^{-54}$ | $4.95 \times 10^{-1}/3.95 \times 10^{-5}/2.52 \times 10^{-4}$ |
| **9 (S1)** | $6.99 \times 10^{-8}/1.73 \times 10^{-25}$ | $1.46 \times 10^{-26}/7.17 \times 10^{-38}/2.94 \times 10^{-43}$ | $4.73 \times 10^{-1}/1.47 \times 10^{-3}/8.14 \times 10^{-6}$ |
| **2 (S2)** | $2.83 \times 10^{-18}/2.93 \times 10^{-25}$ | $1.58 \times 10^{-1}/4.19 \times 10^{-9}/7.44 \times 10^{-8}$ | $7.97 \times 10^{-2}/9.86 \times 10^{-15}/4.73 \times 10^{-11}$ |
| **3 (S2)** | $3.17 \times 10^{-13}/6.08 \times 10^{-18}$ | $8.40 \times 10^{-5}/3.50 \times 10^{-13}/2.28 \times 10^{-7}$ | $6.43 \times 10^{-1}/1.68 \times 10^{-10}/2.76 \times 10^{-10}$ |
| **4 (S2)** | $8.54 \times 10^{-18}/9.18 \times 10^{-26}$ | $4.14 \times 10^{-1}/1.33 \times 10^{-8}/1.36 \times 10^{-5}$ | $6.83 \times 10^{-1}/1.13 \times 10^{-8}/1.48 \times 10^{-5}$ |
| **5 (S2)** | $1.93 \times 10^{-5}/2.50 \times 10^{-14}$ | $3.92 \times 10^{-12}/1.05 \times 10^{-13}/2.41 \times 10^{-18}$ | $5.23 \times 10^{-1}/4.06 \times 10^{-10}/5.57 \times 10^{-5}$ |
| **6 (S2)** | $7.75 \times 10^{-12}/1.14 \times 10^{-21}$ | $9.92 \times 10^{-18}/1.21 \times 10^{-22}/4.57 \times 10^{-24}$ | $3.64 \times 10^{-1}/3.20 \times 10^{-9}/5.76 \times 10^{-8}$ |
| **7 (S2)** | $1.05 \times 10^{-8}/1.39 \times 10^{-20}$ | $1.16 \times 10^{-19}/4.66 \times 10^{-33}/1.30 \times 10^{-34}$ | $5.44 \times 10^{-1}/2.77 \times 10^{-5}/1.69 \times 10^{-8}$ |
| **8 (S2)** | $3.62 \times 10^{-8}/5.09 \times 10^{-43}$ | $6.42 \times 10^{-42}/3.65 \times 10^{-55}/9.24 \times 10^{-54}$ | $7.43 \times 10^{-1}/1.66 \times 10^{-1}/3.28 \times 10^{-1}$ |
| **9 (S2)** | $5.01 \times 10^{-8}/7.66 \times 10^{-24}$ | $4.09 \times 10^{-22}/4.60 \times 10^{-36}/4.36 \times 10^{-42}$ | $6.32 \times 10^{-1}/9.87 \times 10^{-2}/1.23 \times 10^{-3}$ |
| **2 (S3)** | $2.86 \times 10^{-24}/7.75 \times 10^{-28}$ | $6.28 \times 10^{-1}/1.00 \times 10^{-7}/6.50 \times 10^{-4}$ | $8.19 \times 10^{-2}/5.50 \times 10^{-15}/2.69 \times 10^{-2}$ |
| **3 (S3)** | $4.64 \times 10^{-25}/4.33 \times 10^{-20}$ | $2.19 \times 10^{-2}/2.51 \times 10^{-9}/4.71 \times 10^{-5}$ | $3.46 \times 10^{-1}/8.21 \times 10^{-11}/8.31 \times 10^{-4}$ |
| **4 (S3)** | $1.12 \times 10^{-25}/3.25 \times 10^{-26}$ | $6.29 \times 10^{-1}/4.78 \times 10^{-7}/9.24 \times 10^{-3}$ | $9.51 \times 10^{-1}/2.10 \times 10^{-6}/2.42 \times 10^{-2}$ |
| **5 (S3)** | $1.17 \times 10^{-26}/1.18 \times 10^{-13}$ | $8.76 \times 10^{-8}/1.02 \times 10^{-13}/2.64 \times 10^{-16}$ | $5.16 \times 10^{-1}/1.23 \times 10^{-10}/6.93 \times 10^{-7}$ |
| **6 (S3)** | $4.33 \times 10^{-31}/2.19 \times 10^{-20}$ | $2.64 \times 10^{-17}/7.23 \times 10^{-23}/2.82 \times 10^{-23}$ | $2.70 \times 10^{-1}/1.69 \times 10^{-8}/1.48 \times 10^{-6}$ |
| **7 (S3)** | $6.28 \times 10^{-25}/2.68 \times 10^{-17}$ | $6.34 \times 10^{-19}/6.37 \times 10^{-21}/2.15 \times 10^{-27}$ | $3.47 \times 10^{-1}/1.41 \times 10^{-5}/2.61 \times 10^{-3}$ |
| **8 (S3)** | $3.34 \times 10^{-37}/5.18 \times 10^{-31}$ | $1.40 \times 10^{-32}/7.44 \times 10^{-49}/6.80 \times 10^{-52}$ | $7.68 \times 10^{-1}/4.57 \times 10^{-3}/1.24 \times 10^{-5}$ |
| **9 (S3)** | $1.55 \times 10^{-32}/1.65 \times 10^{-17}$ | $3.45 \times 10^{-23}/1.20 \times 10^{-33}/1.35 \times 10^{-41}$ | $8.45 \times 10^{-1}/8.27 \times 10^{-3}/4.64 \times 10^{-8}$ |
| **Mean (S1)** | $1.43 \times 10^{-71}/2.52 \times 10^{-150}$ | $3.60 \times 10^{-81}/3.44 \times 10^{-133}/3.17 \times 10^{-108}$ | $2.84 \times 10^{-1}/1.47 \times 10^{-56}/1.37 \times 10^{-42}$ |
| **Mean (S2)** | $1.74 \times 10^{-67}/7.56 \times 10^{-130}$ | $2.23 \times 10^{-47}/7.96 \times 10^{-124}/3.33 \times 10^{-111}$ | $6.95 \times 10^{-1}/1.21 \times 10^{-44}/3.44 \times 10^{-36}$ |
| **Mean (S3)** | $1.17 \times 10^{-124}/5.80 \times 10^{-118}$ | $3.67 \times 10^{-29}/2.97 \times 10^{-103}/3.00 \times 10^{-85}$ | $7.16 \times 10^{-1}/1.76 \times 10^{-46}/3.40 \times 10^{-24}$ |

**Supplementary Tab. 3 | *P* values for the one-shot classification experiment (Fig. 5b and Supplementary Fig. S3; n = 100 per task)**. The first column represents the number of abstract concepts. We applied the two-tailed paired *t*-test for statistical analysis.

| **Abstract concepts (Number)** | **Including concrete concepts** |
|---|---|
| living thing, non-living thing (2) | all 50 concrete concepts |
| invertebrate, vertebrate, non-living thing (3) | all 50 concrete concepts |
| bird, invertebrate, mammal, artifact (4) | 48 concrete concepts<br>(except common iguana, goldfish) |
| living thing, instrumentality, covering, home appliance, structure (5) | all 50 concrete concepts |
| instrumentality, covering, vertebrate, invertebrate, structure, home appliance (6) | all 50 concrete concepts |
| instrumentality, vertebrate, clothing, invertebrate, structure, protective covering, home appliance (7) | all 50 concrete concepts |
| arthropod, bird, ungulate, home appliance, equipment, musical instrument, vehicle, clothing (8) | 29 concrete concepts<br>(see Supplementary Fig. S1) |
| structure, container, home appliance, vertebrate, covering, craft, invertebrate, device, equipment (9) | 47 concrete concepts<br>(except cannon, hammock, pincer) |

**Supplementary Tab. 4 | Concept combinations across eight abstraction levels.** The different combinations of abstract concepts were used in one-shot abstract classification tasks (Fig. 5b and Supplementary Fig. S4). These abstract concepts are derived from hypernyms (superclasses) of WordNet for 50 concrete concepts.

## Supplementary Text

### Implementation details of brain-in-the-loop supervised learning (Fig. 2)

Here, we provide details on the architecture configuration and hyperparameter settings of the brain-in-the-loop supervision approach.

The brain-in-the-loop supervised learning framework can integrate with any pre-trained foundation model. Models leveraged in this work include **1)** OpenAI trained CLIP-base (ViT-B/16 backbones) and CLIP-large (ViT-L/14 backbones)[1]; **2)** Meta AI trained SimCLR-small (ViT-S/16 backbones), SimCLR-base (ViT-B/16 backbones) and SimCLR-large (ViT-L/16 backbones)[2]; **3)** Meta AI trained DINOv2-small (ViT-S/14 backbones), DINOv2-base (ViT-B/14 backbones) and DINOv2-large (ViT-L/14 backbones)[3]. These pre-trained model parameters can be available from their official implementation on GitHub and HuggingFace. We used voxels from brain higher visual areas including the lateral occipital complex (LOC, with voxel counts of 3,072 for S1; 2,627 for S2; 3,745 for S3), fusiform face area (FFA, with voxel counts of 2,874 for S1; 1,574 for S2; 980 for S3), and parahippocampal place area (PPA, with voxel counts of 1,157 for S1; 1,419 for S2; 908 for S3) for brain-in-the-loop supervised learning.

During the training phase, the framework consists of three learnable sub-models: an image encoder (MLP), an fMRI encoder (ViT[4]), and a shared GNN. Specifically, the image encoder consists of an input layer (with the ReLU activation[5] and an embedding size of 256) and an output linear layer. The fMRI encoder is a standard ViT (the embedding dimension matches the size of the input image embeddings) with 12 layers and 8-head self-attention. The shared GNN is a multiplex attention graph neural network[6–8] with 3 layers of 4-head self-attention. Our training data consists of 206,848 object images (drawn from ImageNet-21K) and 6,000 fMRI signals, covering a total of 150 categories. The remaining 50 categories in the DIR dataset constitute the test set. We adopted the same data augmentation strategy as in BrainSem[9] and MindArt[10]. During the putative correspondence construction phase, we run Sinkhorn algorithm with 100 iterations and set $K_1 = K_2 = 20$ for the Gumbel sampler. The framework is implemented in PyTorch, and we train it using the Adam optimizer[11] (learning rate is 2e-5, batch size is 256) with parameters $\beta_1 = 0.9$ and $\beta_2 = 0.999$.

### Implementation details of text reconstruction model for the conceptual manifold analysis (Fig. 6)

In this section, we provide details on the architecture configuration and hyperparameter settings of the text reconstruction model, which is used in our cognitive generalization analysis experiments.

**Implementation details.** To generate descriptive sentences from the conceptual manifold, we built a text reconstruction architecture based on GPT-2[12] (we use the pre-trained GPT-$2_{\text{Base}}$ model available on HuggingFace). Given an image [CLS] token, we first apply a trainable linear projection to reshape it into a sequence of patch embeddings with a size of $8 \times 768$. Since the image encoders and GPT-2 operate in different latent spaces, we employ a set of cross-attention layers to bridge them. Following Ramos *et al.*[13], each of the 12 layers of the GPT-2 decoder incorporates a 12-head cross-attention layer, and we reduce the default dimensionality (64) of the projection matrices in the attention layers to 16. We train the text reconstruction model on the COCO dataset with standard Karpathy splits by minimizing the cross-entropy loss with AdamW optimizer[14] (learning rate is 1e-4, batch size is 256, $\beta_1 = 0.9$ and $\beta_2 = 0.999$). Note that we train two text reconstruction models for CLIP—one with and one without brain-in-the-loop supervision—using the same training setup, including architecture configuration and learning hyperparameters. The reconstruction model is implemented in PyTorch and trained for 20 epochs on four NVIDIA GeForce RTX 3090 GPUs.

**Loss function.** Given image-text pairs $(x, t)$ (drawn from the COCO dataset[15]), the loss function used to train the GPT-2[12]-based text decoder is defined as follows:

$$\mathcal{L}_{text} = -\sum_{l=1}^{M} logP(t_i | t_{<i}, \text{linear}(\text{DNN}(x); \Theta)$$

where $P(t_i | t_{<i}, \text{linear}(\text{DNN}(x))$ denotes the likelihood of predicting the next token, conditioned on previous tokens and the output of the linear layer, $M$ denotes the length of the target token sequence, and $\Theta$ is the parameters of cross-attention[16] modules and the linear layer.

**Computation of the representational dissimilarity matrix (Fig. 1)**

To compute representational dissimilarity matrices (RDMs) for humans, we used cosine distance, i.e., 1 - *r* (where *r* denotes cosine similarity) to measure the dissimilarity between BOLD response patterns (n = 50). Each element in the human RDMs was computed using the first 8 principal components derived from all voxels in the HVC, and then averaged across all subjects (voxel number: n = 7,103 for S1; 5,620 for S2; 5,633 for S3). It should be noted that the model RDMs were computed in a similar manner by using the latent embeddings of 50 ground truth image stimuli.

**Similarity between conceptual hierarchy structures (Fig. 4)**

To quantify the semantic structural similarity between DNNs and humans, we employed human-defined semantic network, namely, WordNet[17], a large conceptual-semantic relations database, as a validation tool. Specifically, we calculated the pairwise shortest paths among 50 synsets as ground truth concept similarity matrix (CSM). The rows and columns of the matrix are ordered by the average-linkage hierarchical clustering algorithm[18] (implemented in SciPy, with default parameters) to capture and expose the semantic structure patterns in the WordNet. For the DNN models, we compute the pairwise cosine distances between 50 concept embeddings to construct the CSMs, where each concept embedding is obtained by averaging the image representations within the same category (n = 66,067, 275–2,115 images per category). Note that the rows and columns of the CSMs for the DNN models are sorted according to the WordNet clustering results (see Supplementary Tab. 1 for the sorted row/column labels).

**The triplet odd-one-out experiments (Fig. 4)**

The triplet odd-one-out task is a cognitive and psychological test used to investigate how humans understand and categorize concepts. Below we provide details on the experiment settings.

**Dataset.** Human odd-one-out responses used in our research originated from THINGS[19], a large human behavioral database of 4.7 million unique human similarity judgments for 1,854 object images from 12,340 participants. The THINGS dataset is widely used to assess human notions of object similarity. For each task, participants are presented with three object images and asked to select the image that is least similar to the other two. The presented images depict objects commonly encountered in daily life, and it can be assumed the most participants are sufficiently familiar with these objects to name them.

To assess the degree of alignment between humans and large models in object similarity judgment tasks, we first feed 1,854 object images from the THINGS object concept and image dataset[20] into the models to extract latent embeddings. Next, given a triplet odd-one-out task $(\alpha_i, \alpha_j, \alpha_k)$ in the THINGS database[19], we construct a similarity matrix $X \in \mathbb{R}^{3\times3}$, where $X_{i,j} \coloneqq x_i^{\mathrm{T}} x_j$ is the scalar product between model embeddings of images $\alpha_i$ and $\alpha_j$. We then determine the odd-one-out by excluding the two most similar image embeddings, i.e., $arg\,max(X_{i,j}), i \neq j$. Subsequently, odd-one-out accuracy is defined as the proportion of correct predictions made by the model out of all predictions, with the ground truth labels being human responses collected from the THINGS database[19]. In our experiment, we performed n = 100 independent trials, each time sampling 10,000 human responses without replacement from the THINGS odd-one-out data, following a uniform distribution.

**Settings of the experimental tasks (Fig. 5)**

Here, we give the details of the complex tasks employed in this study. All the comparative tests were done with the same training setup including the model configuration and learning hyperparameters.

**One-shot classification.** This task comprises two phases: the training phase, which aims to learn a classification model from only a few labeled examples, and the adaptation phase to evaluate the model's generalization ability[21]. In the training phase, the multinomial logistic regression algorithm and the cross-entropy loss with an $L_2$ penalty term are used to train a linear classifier. Here the training samples were randomly selected. In the adaptation phase, we measured the classification accuracy of the linear model on test images including 50 unseen concepts (100 independent trials per classification task), all conducted using the same hyperparameters, model, and training setup. See Supplementary Tab. 4 for the different abstraction levels involved in the one-shot classification task.

**Generalized OOD recognition.** Our tasks define 31 recognition targets (see Supplementary Tab. 2 for details). For each recognition target, we used two concrete classes (5 images per category) as the training set and the remaining 48 concrete classes (unknown) were used for testing. Similar to the above, we fitted a linear model by using logistic regression. The model is then used to detect whether other categories belong to this target (chance level = 50%). Since test samples do not fall into any known class, the test set can be considered as OOD with a semantic shift. We repeated this process 100 times for each recognition task.

**Brain-to-image and image-to-image retrieval.** The objective of brain-to-image retrieval is to decode stimulus semantics from brain activity measured using fMRI[22]. To this end, we extracted [CLS] embeddings of all real stimulus images using CLIP-base model. Next, we predicted the visual features of the stimulus images from brain signals using the trained fMRI encoder. Finally, the predicted fMRI representations were used to identify the ground truth stimuli (n = 50) by calculating the cosine similarity between embeddings. We used 500-trial n-way ($2 \leq n \leq 50$) top-1 accuracy to evaluate the decoding performance, where "500-trial" indicates the identification task was performed 500 times, and "n-way" means accuracies were measured in n-1 randomly selected stimulus classes plus the ground truth. Moreover, for the image-to-image retrieval task, we extracted [CLS] embeddings of test images (n = 66,067) using the CLIP-base and brain-in-the-loop supervised CLIP-base models respectively, and then ranked the test images based on their similarity to the given query embedding. Notably, unlike traditional retrieval tasks, here we enforced a constraint that the top-k retrieved images must belong to different categories.

**Analysis of cognitive generalization on the conceptual manifold (Fig. 6)**

Here we present the details of the conceptual manifold analysis experiments. Our hypothesis is that brain-in-the-loop supervision produces a more efficient low-dimensional conceptual manifold at cognitive-level, e.g., the low-dimensional latent space captures typical judgements of concepts, allowing us to accurately describe and rate objects along these dimensions[23].

To validate this, we first trained two identical decoders (with same training strategies) on the COCO dataset[15] to translate image latent embeddings to natural language—both before and after brain-in-the-loop supervision (Supplementary Fig. S9a). Specifically, given an image [CLS] embedding, we first fed it into a linear projection layer to generate a sequence of continuous representations ($n = 8$). Then, a cross-attention[16] module was leveraged to connect DNNs and the GPT-2 model, where the parameters of GPT-2 are frozen, and only the attention module is trainable (see Supplementary for details of the architecture configuration and training procedures). Our goal is not to generate image captions, but rather to reconstruct semantics from any position on the low-dimensional manifold to assess the cognitive coherence.

To reconstruct semantics from low-dimensional spaces, we generated a PCA-based manifold from 13,750 test images (50 unseen categories, 275 image per category). This practice allows us to generate arbitrary samples from the first two principal components and map them back into the original high-dimensional embedding spaces using PCA inverse transform operation. Subsequently, the reconstructed embeddings can be fed into the decoders to be translated into text (Supplementary Fig. S9b). Note that these object images are drawn from ImageNet-21K rather than the COCO dataset, which ensures that the test samples are out-of-distribution (OOD) for the semantic decoder.

In the semantic consistency analysis, we conducted $n = 100$ independent trials, each time sampling 1,000 2-D data points following a uniform distribution in the range [-5, 5] (for both the first and second principal components). We used the nearest neighbors in real samples as ground truth labels to assess the accuracy of reconstructed semantics of each sampled point. Namely, consistency accuracy is defined as the proportion of correctly identifying the ground-truth label from all candidate labels ($n = 8$ and $n = 50$). We employed cosine distance to measure the semantic similarity between reconstructed text and candidate labels. Note that, in the summarization and arithmetic experiments, reconstruction operations were performed on the original high-dimensional embedding, rather than the PCA-based space (Supplementary Fig. S9c).